\documentclass[letterpaper]{article} % DO NOT CHANGE THIS
\usepackage{aaai25}  % DO NOT CHANGE THIS
\usepackage{times}  % DO NOT CHANGE THIS
\usepackage{helvet}  % DO NOT CHANGE THIS
\usepackage{courier}  % DO NOT CHANGE THIS
\usepackage[hyphens]{url}  % DO NOT CHANGE THIS
\usepackage{graphicx} % DO NOT CHANGE THIS
\usepackage{natbib}  % DO NOT CHANGE THIS AND DO NOT ADD ANY OPTIONS TO IT
\usepackage{caption} % DO NOT CHANGE THIS AND DO NOT ADD ANY OPTIONS TO IT
\usepackage{times}
\usepackage{latexsym}
\usepackage{amssymb}
\usepackage{color}
\usepackage{pifont}
\newtheorem{theorem}{Theorem}
\newtheorem{proof}{Proof}
\usepackage{amsmath}
\usepackage{booktabs}
\usepackage{algorithm}
\usepackage{algorithmic}
\usepackage{tcolorbox}
\usepackage{multirow}
\usepackage{colortbl}
\usepackage{graphicx}

\usepackage{newfloat}
\usepackage{listings}
\DeclareCaptionStyle{ruled}{labelfont=normalfont,labelsep=colon,strut=off} % DO NOT CHANGE THIS
\floatstyle{ruled}
\newfloat{listing}{tb}{lst}{}
\floatname{listing}{Listing}
\title{Walk Wisely on Graph: Knowledge Graph Reasoning  with Dual Agents via Efficient Guidance-Exploration}
\author{
    Zijian Wang\textsuperscript{\rm 1,2},
    Bin Wang\textsuperscript{\rm 1}\thanks{Bin Wang is the corresponding author.},
    Haifeng Jing\textsuperscript{\rm 1,\rm 3},
    Huayu Li\textsuperscript{\rm 1},
    Hongbo Dou\textsuperscript{\rm 1}
}

\affiliations{
    \textsuperscript{\rm 1}College of Computer Science and Technology \& Qingdao Institute of Software, China University of Petroleum(East China), China\\
    \textsuperscript{\rm 2}College of Science, China University of Petroleum(East China), China\\
    \textsuperscript{\rm 3}School of Software \& Microelectronics, Peking University, China 
    alococo710@gmail.com, \{wangbin,lhyzj\}@upc.edu.cn, 2201120005@stu.pku.edu.cn, hongboDou@163.com
}

\usepackage{bibentry}
\begin{document}

\maketitle

\begin{abstract}
Recent years, multi-hop reasoning has been widely studied for knowledge graph (KG) reasoning due to its efficacy and interpretability. However, previous multi-hop reasoning approaches are subject to two primary shortcomings. First, agents struggle to learn effective and robust policies at the early phase due to sparse rewards. Second, these approaches often falter on specific datasets like sparse knowledge graphs, where agents are required to traverse lengthy reasoning paths. To address these problems, we propose a multi-hop reasoning model with dual agents based on hierarchical reinforcement learning (HRL), which is named \textbf{FULORA}. FULORA tackles the above reasoning challenges by e\textbf{F}ficient G\textbf{U}idance-Exp\textbf{LORA}tion between dual agents. The high-level agent walks on the simplified knowledge graph to provide stage-wise hints for the low-level agent walking on the original knowledge graph. In this framework, the low-level agent optimizes a value function that balances two objectives: (1) maximizing return, and (2) integrating efficient guidance from the high-level agent. Experiments conducted on three real-word knowledge graph datasets demonstrate that FULORA outperforms RL-based baselines, especially in the case of long-distance reasoning.
\end{abstract}

\section{Introduction}
Knowledge graphs (KGs) are designed to represent the world knowledge in a structured way. There are various downstream NLP tasks especially knowledge-driven services, such as query answering \cite{guu2015traversing, DBLP:journals/corr/abs-1903-02419}, relation extraction \cite{DBLP:conf/acl/MintzBSJ09,DBLP:conf/tal/ReiplingerWK14} and dialogue generation \cite{DBLP:conf/acl/HeBEL17}. However, a significant proportion of KGs are severely incomplete, which constrains their efficacy in numerous tasks. Consequently, this study concentrates on automatic knowledge graph (KG) reasoning, also as known knowledge graph completion (KGC). 

Over recent years, embedding-based models \cite{DBLP:conf/nips/BordesUGWY13,DBLP:conf/aaai/LinLSLZ15} have effectively preserved KG structural information for single-hop reasoning but lack interpretability. To address this, reinforcement learning (RL) frameworks \cite{DBLP:conf/emnlp/XiongHW17,DBLP:conf/iclr/DasDZVDKSM18} have been introduced to compose single-hop triplets into multi-hop reasoning chains. Recent advances in deep learning (DL) and RL \cite{DBLP:conf/emnlp/WangLPM19,DBLP:conf/emnlp/LvHHLLZZKW20,DBLP:conf/asunam/NikopensiusMPS23,wang2025windows} further enhance multi-hop reasoning. For instance, AttnPath \cite{DBLP:conf/emnlp/WangLPM19} employs attention mechanisms to guide agents, preventing them from stalling at the same node.

\begin{figure}[t] \label{fig1}
    \centering
    \includegraphics[width=1\linewidth]{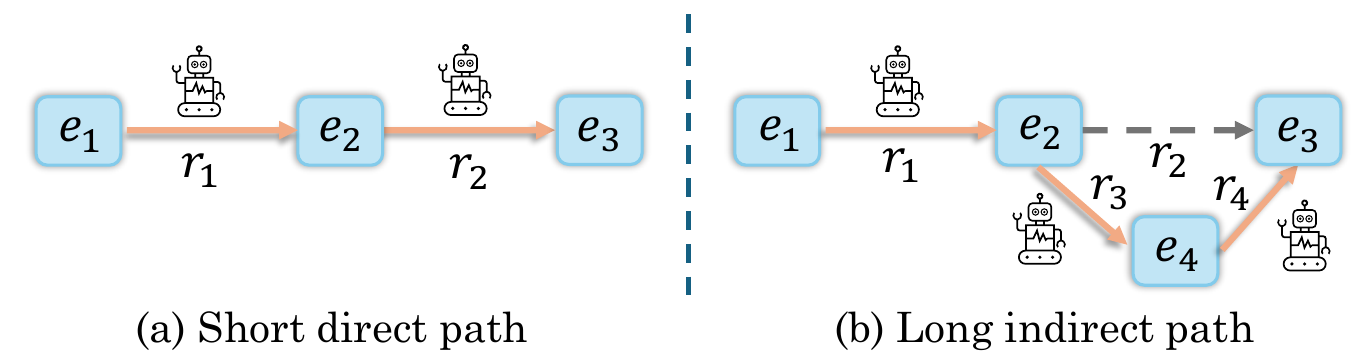}
    \caption{An illustrative example of short direct path and long indirect path. When no short direct path exists, the agent searches for a long indirect path.}
    \label{fig:enter-label}
\end{figure}

Although existing multi-hop reasoning models have achieved impressive results, there is a noteworthy issue with these models that they only perform well when the agent's reasoning path length is short. The drawback that the agent relies heavily on short reasoning chains is fatal in certain datasets, such as sparse KG. Figure 1 is an illustration of short direct path and long indirect path. It's challenging for an agent to identify a short direct path due to the sparsity which makes the relation $r_2$ do not exist. The optimal path for agent is $e_1 \rightarrow e_2 \rightarrow e_4 \rightarrow e_3$ we called long indirect path instead of $e_1 \rightarrow e_2 \rightarrow e_3$. From this point, enhancing the agent's long-distance reasoning ability is one solution to alleviate the multi-hop models' poor performance in the sparse KG. 

However, the dimension of the discrete action space at each space is large \cite{DBLP:conf/iclr/DasDZVDKSM18}. As the path length increases, the choices the agent faced will grow exponentially. The most relevant to our work is CURL \cite{DBLP:conf/aaai/ZhangY0L022} which proposed a dual-agent framework and mutual reinforcement rewards to assign one of the agents (GIANT) searching on cluster-level paths quickly and providing stage-wise hints for another agent (DWARF). Compared to the classic multi-hop model MINERVA \cite{DBLP:conf/iclr/DasDZVDKSM18}, CURL does improve long-distance reasoning ability. But CURL also has two drawbacks, which make CURL's performance inconsistent: (1) Mutual reinforcement reward mechanism forces the low-level agent DWARF to adopt similar policies to the high-level agent GIANT, even if GIANT's policies is not well enough. This may result in false-negative rewards to the intermediate actions which are reasonable. (2) At the early phase, DWARF and GIANT adopt a near-random policy which makes low training efficacy. For briefness, we call the high-level agent as GIANT and the low-level agent as DWARF like \cite{DBLP:conf/emnlp/BaiLL0QDX22}.

In light of these challenges, we propose FULORA, a robust dual-agent framework for KG reasoning with taking full advantage of entity embedding and relation embedding in the KG. FULORA seamlessly makes the self-exploration and path-reliance trade-off of DWARF through a supervised learning method. This unique mechanism will enable DWARF to make its own decisions while receiving meaningful guidance from GIANT, instead of relying entirely on the command of GIANT. Moreover, with the aim to make better use of the entity embedding and relation embedding, FULORA introduces attention mechanism and dynamic path feedback to DWARF and GIANT respectively. Intuitively, GIANT searches a feasible path as soon as possible to guide DWARF to reduce the search space, while DWARF adopts diversified exploration policies in the constrictive search space to prevent over-dependence on GIANT. In this way, DWARF walking on the original KG has both excellent long distance reasoning ability and short direct path utilization ability.

\section{Related Work}
In line with the focus of our work, we provide a brief overview of the background and related work on knowledge graph reasoning. 
\subsection{Knowledge Graph Embedding}
Knowledge graph embedding (KGE) methods map entities to vectors in low-dimensional embedding space, and model relations as transformations between entity embeddings \cite{DBLP:conf/emnlp/BaiLL0QDX22}. Once we map them to low-dimensional dense vector space, we can use modeling methods to perform calculations and reasoning \cite{DBLP:journals/ijon/MaSHC18}. Prominent examples include TransE \cite{DBLP:conf/nips/BordesUGWY13}, TransR \cite{DBLP:conf/aaai/LinLSLZ15}, ConvE \cite{DBLP:conf/aaai/DettmersMS018}, TuckER \cite{DBLP:conf/emnlp/BalazevicAH19} and LMKE \cite{DBLP:conf/ijcai/WangHLX22}, which are equipped with a scoring function that maps any triplet $(e_s,r_q,e_o)$ to a scalar score. While KGE is effective at capturing simpler relationships in the graph, such as first-order adjacency relationships, it struggles with more complex reasoning tasks, particularly those involving multi-hop reasoning \cite{yao2019kg,wang2023mega}.

\subsection{GNN-based Reasoning}
Graph neural networks (GNNs) \cite{scarselli2008graph, velivckovic2017graph, xu2018powerful} are a class of models used for representation learning, specifically designed to encode the structural information of graphs. In the context of link prediction, common frameworks often rely on an auto-encoder formulation, where GNNs generate node embeddings, and edges are predicted as a function of node pairs. These frameworks are inductive when node features are provided, but they become transductive when such features are not available. Another set of frameworks, including SEAL \cite{zhang2018link} and GraIL \cite{teru2020inductive}, explicitly encode the subgraph surrounding each node pair for link prediction. Recent advancements in this field include works such as NBFNet \cite{zhu2021neural} and RED-GNN \cite{zhang2022knowledge}. The former solves the path formulation with learned operators in the generalized Bellman-Ford algorithm while the latter makes use of dynamic programming to recursively encodes multiple r-digraphs with shared edges, and utilizes query-dependent attention mechanism to select the strongly correlated edges. However, these methods primarily focus on the structure of the knowledge graph (KG) itself, specifically the graph structure information of the central node, without addressing how to enhance the ability for long-distance reasoning.

\subsection{Multi-hop Reasoning}
The advancement of deep reinforcement learning (DRL) has sparked interest in applying DRL to path-finding tasks. The first significant work combining DRL and KG reasoning is DeepPath \cite{DBLP:conf/emnlp/XiongHW17}, which inspired subsequent models, though it requires prior knowledge of the target entity. In contrast, MINERVA \cite{DBLP:conf/iclr/DasDZVDKSM18} eliminates this requirement, allowing the agent to traverse the knowledge graph until it finds the target. Recent developments have explored the use of powerful neural networks for generating walking policies. M-Walk \cite{DBLP:conf/nips/ShenCHGG18} uses an RNN \cite{elman1990finding} to record agent trajectories and optimize rewards with a Monte Carlo Tree Search (MCTS) \cite{coulom2006efficient}. GRL \cite{DBLP:journals/kbs/WangJHC20} combines GAN \cite{goodfellow2014generative} and LSTM \cite{hochreiter1997long} to generate new trajectory sequences, enabling the agent to reason not only within the original graph but also in automatically generated sub-graphs, extending relational paths until target entities are found. HMLS \cite{DBLP:journals/tkde/ZhengCWZYZ24} improves the generalizability and effectiveness of multi-hop reasoning in few-shot scenarios by exploiting hard relations and hierarchical relation structures.  While these models perform well in short-distance reasoning (path length = 3), their performance in long-distance reasoning tasks remains suboptimal. Unlike previous multi-hop reasoning methods, FULORA focuses more on the agents' long-distance reasoning ability and the significant impact of sparse rewards on training efficiency.
\begin{figure*}[t] \label{fig2}
    \centering
    \includegraphics[width=0.95\textwidth]{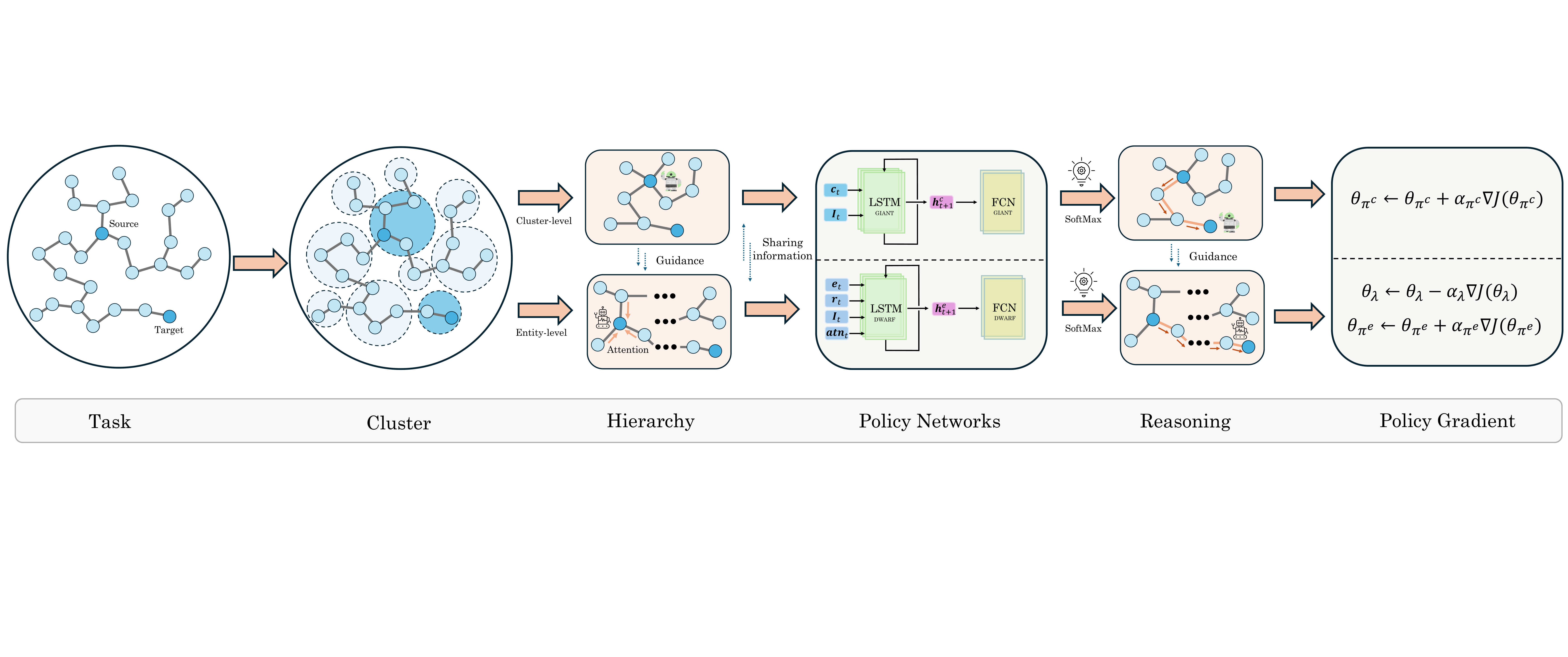}
    \caption{An overview of FULORA framework. \ding{182} Given a KG $\mathcal{G}$, We first pre-embed the KG using TransE and then apply K-means clustering to generate the cluster-level KG $\mathcal{G}^c$. \ding{183} We design separate policy networks for GIANT and DWARF, using cluster-level KG $\mathcal{G}^c$ and entity-level KG $\mathcal{G}^e$ as inputs. The hidden states of GIANT and DWARF $\mathbf{h}_t^c$ and $\mathbf{h}_t^e$ share information, facilitating communication. \ding{184} To enable DWARF to better leverage the KG structure, we apply the graph attention mechanism and feed the resulting attention vector into the policy network. Dynamic Path Feedback alleviates the near-random policy issue caused by sparse rewards in early training phase, allowing GIANT to provide high-quality guidance to DWARF sooner.
 }
    \label{fig:enter-label}
\end{figure*}

\section{Preliminary}
In this section, we commence with the problem definition of our work, followed by the introduction of environment representation.
\subsection{Problem Definition}
We formally define the research problem of this paper in the following part. The knowledge graph is defined as a directed graph $\mathcal{G} = \{\mathcal{E},\mathcal{R}\}$, where $\mathcal{E}$ is a set of all entities and $\mathcal{R}$ is a set of all relations. A link triplet $l$ consists of the source entity $e_s \in \mathcal{E}$, the target entity $e_o \in \mathcal{E}$ and the relation $r \in \mathcal{R}$, i.e., $l = (e_s,r,e_o)$. In the real world, a link triplet corresponds to a fact tuple. For instance, we can represent the fact \textbf{whale is a mammal} as $(whale,is \ a,mammal)$. We follow the definition in prior graph walking models, that is, query answering \cite{DBLP:conf/iclr/DasDZVDKSM18,DBLP:conf/aaai/ZhangY0L022}. In the applications such as searching and query answering, most problems are to infer another entity when we only know the source entity $e_s$ and the query relation $r_q$, which can be formed by an incomplete link triplet $l_m = (e_s,r_q,?)$, where "?" indicates the target entity $e_o$ is unknown and needs to be found in the KG. 
\subsection{Environment Representation}

\textbf{State.} The state consists of two entities—current entity \( e_t \) and source entity \( e_s \)—and the query relation. The current entity \( e_t \) is state-dependent, reflecting the agent's reasoning condition, while the source entity \( e_s \) and query relation are shared globally. Formally, \( s_t = (e_t, e_s, r_q) \in \mathcal{S} \).
\\
\textbf{Action.} The agent interacts with the environment by selecting an action \( a_t \in \mathcal{A} \), corresponding to an outgoing edge of the graph \( \mathcal{G} \). Formally, \( a_t = \{(r_{t+1}, e_{t+1}) | (e_t, r_{t+1}, e_{t+1}) \in \mathcal{G}\} \). To allow termination, the agent can also stay at the current entity by selecting a self-loop edge.
\\
\textbf{Transition.} A state transition occurs when the agent takes an action. The transition function \( \delta: \mathcal{S} \times \mathcal{A} \rightarrow \mathcal{S} \) is defined as \( \delta(s_t, a_t) \), which produces a new state. Multi-hop reasoning typically selects a neighboring entity randomly, as entities often have multiple neighbors connected by the same relation \cite{DBLP:conf/emnlp/XiongHW17, DBLP:conf/iclr/DasDZVDKSM18}.
\\
\textbf{Reward.} After the agent takes an action, if the corresponding entity is the correct target, it receives a reward of 1; otherwise, it receives a reward of 0.

\section{Methodology}
In this section, we propose FULORA, an extended dual-agent framework for knowledge graph reasoning via efficient guidance-exploration. As illustrated in Figure 2, FULORA is able to improve the exploration efficiency of GIANT via dynamic feedback mechanism, so as to provide more reliable guidance for DWARF. In a corresponding manner, DWARF employs a supervised learning approach to achieve a balance between guidance and exploration. Subsequently, it aggregates messages emitted by all neighbors of the current entity via an attention mechanism. Next, we will delve into the specifics of each aforementioned components.

\subsection{Embedding Generation}
Consistent with \cite{DBLP:conf/aaai/ZhangY0L022}, we employ TransE \cite{DBLP:conf/nips/BordesUGWY13} due to its efficiency in encoding the structural proximity information of KG to generate pre-trained entity embeddings, then we divide the original KG into $N$ clusters by utilizing K-means. In order for GIANT can easily walk on the cluster-level graph $\mathcal{G}^c$ while preserving the relation information of the original KG $\mathcal{G}$, we add a link to two clusters if there is at least one entity-level edge between them, which is detailed in Appendix E.

\subsection{Policy Networks}
In our model, we utilize a three-layer LSTM, enabling the agent to memorize and learn from the actions taken before. In contrast to previous models, the necessity arises to design the network separately in this case, given that two agents are walking on the KG. An agent based on LSTM encodes the recursive sequence as a continuous vector $\mathbf{h}_t \in \mathbb{R}^{2d}$. Specifically, the hidden state embedding of GIANT is $\mathbf{h}^c_t$ while the hidden state embedding of DWARF is $\mathbf{h}^e_t$. Their initial hidden state is $\mathbf{0}$. In addition, we define an information sharing vector $\mathbf{I}_t = [\mathbf{h}^e_t;\mathbf{h}^c_t]$ for GIANT and DWARF to share path information. To a certain extent, cluster-level paths are complementary to entity-level paths, as they ensure the sharing of essential path information from GIANT to DWARF. \\
\textbf{For GIANT.}\quad We denote the current cluster embedding at time step $t$ by $\mathbf{c}_t \in \mathbb{R}^{2d}$. The action representation $\mathbf{a}_t^c$ is given by the cluster embedding itself, i.e., $\mathbf{a}_t^c = \mathbf{c}_t \in \mathbb{R}^{2d}$ because the action corresponds to the next outgoing cluster. The history embedding is updated according to LSTM dynamics:
\begin{align}
    & \mathbf{h}_t^c = \text{LSTM}_c(\mathbf{W}^c[\mathbf{h}_{t-1}^c;\mathbf{I}_{t-1}],\mathbf{a}_{t-1}^c),
\end{align}
where $\mathbf{W}^c \in \mathbb{R}^{2d\times6d}$ is a projection matrix to maintain shape. \\
\textbf{For DWARF.}\quad Unlike GIANT, which operates on cluster-level KGs, DWARF faces greater challenges on the original KG. Entities often have multiple aspects. For instance, a professor may have both professional relations (e.g., \textit{worksForUniversity}) and family relations (e.g., \textit{spouse}). Additionally, cluster-level paths are typically shorter than entity-level paths, requiring DWARF to have enhanced long-distance reasoning. Consequently, DWARF should prioritize relations and neighbors most relevant to the query. Therefore, we integrate the Graph Attention mechanism \cite{DBLP:journals/corr/abs-1710-10903} into DWARF. We adopt the same approach as AttnPath \cite{DBLP:conf/emnlp/WangLPM19} to obtain the attention vector $\mathbf{atn}_{t-1}$. DWARF's history embedding $\mathbf{h}_t^e$ can be obtained from 
\begin{equation}
    \mathbf{h}_t^e = \text{LSTM}_e(\mathbf{W}^e[\mathbf{h}_{t-1}^e;\mathbf{atn}_{t-1};\mathbf{I}_{t-1}],\mathbf{a}_{t-1}^e),
\end{equation}
where $\mathbf{W}^e \in \mathbb{R}^{2d\times7d}$ is a projection matrix, while the action representation $\mathbf{a}_{t}^e$ is the concatenation of the relation embedding $\mathbf{r}_t \in \mathbb{R}^d$ and the end node embedding $\mathbf{e}_t \in \mathbb{R}^d$, i.e., $\mathbf{a}_{t}^e = [\mathbf{r}_t;\mathbf{e}_t] \in \mathbb{R}^{2d}$. 

\noindent
\textbf{Policy Generation.} \quad To predict the next cluster for GIANT and the next entity for DWARF, we apply a two-layer feedforward network on the concatenation of their last LSTM states and current RL state embeddings,
\begin{equation}
    \begin{aligned}
           & \mathbf{d}^c_t = \text{SoftMax}(\mathbf{A}_t^c \times \mathbf{W}_2^c \text{ReLU}(\mathbf{W}^c_1[\mathbf{c}_t;\mathbf{h}_t^c])), \\
           & a_t^c \sim \text{Categorical}(\textbf{d}^c_t),
    \end{aligned}
\end{equation}
\begin{equation}
    \begin{aligned}
        & \mathbf{d}^e_t = \text{SoftMax}(\mathbf{A}_t^e \times \mathbf{W}_2^e \text{ReLU}(\mathbf{W}^e_1[\mathbf{e}_t;\mathbf{r}_q;\mathbf{h}_t^e])), \\
        & a_t^e \sim \text{Categorical}(\textbf{d}^e_t),
    \end{aligned}
\end{equation}
where $\mathbf{W}_1^c \in \mathbb{R}^{4d\times4d}$, $\mathbf{W}_2^c \in \mathbb{R}^{2d\times4d}$, $\mathbf{W}_1^e \in \mathbb{R}^{4d\times4d}$, $\mathbf{W}_2^e \in \mathbb{R}^{2d\times4d}$ are the matrices of learnable weights to maintain dimension of history embedding. While $\mathbf{A}_t^c \in \mathbb{R}^{|A_t^c|\times2d},\mathbf{A}_t^e \in \mathbb{R}^{|A_t^e|\times2d}$ represent the embeddings of all next possible actions for GIANT and DWARF respectively.

\subsection{Efficient Guidance-Exploration}
As mentioned above, to address the issue of sparse rewards and a large action space, we aim for GIANT to guide DWARF in reducing the action space. However, GIANT's guidance is not always beneficial due to two key issues: 
\begin{itemize}
    \item \textbf{Poor guidance.} Poor guidance can lead DWARF to incorrect answers.
    \item \textbf{Policy shift.} A policy which is suitable for GIANT may not be fully applicable to DWARF.
\end{itemize}

\begin{figure} \label{fig3}
    \centering  
    \includegraphics[width=0.5\textwidth]{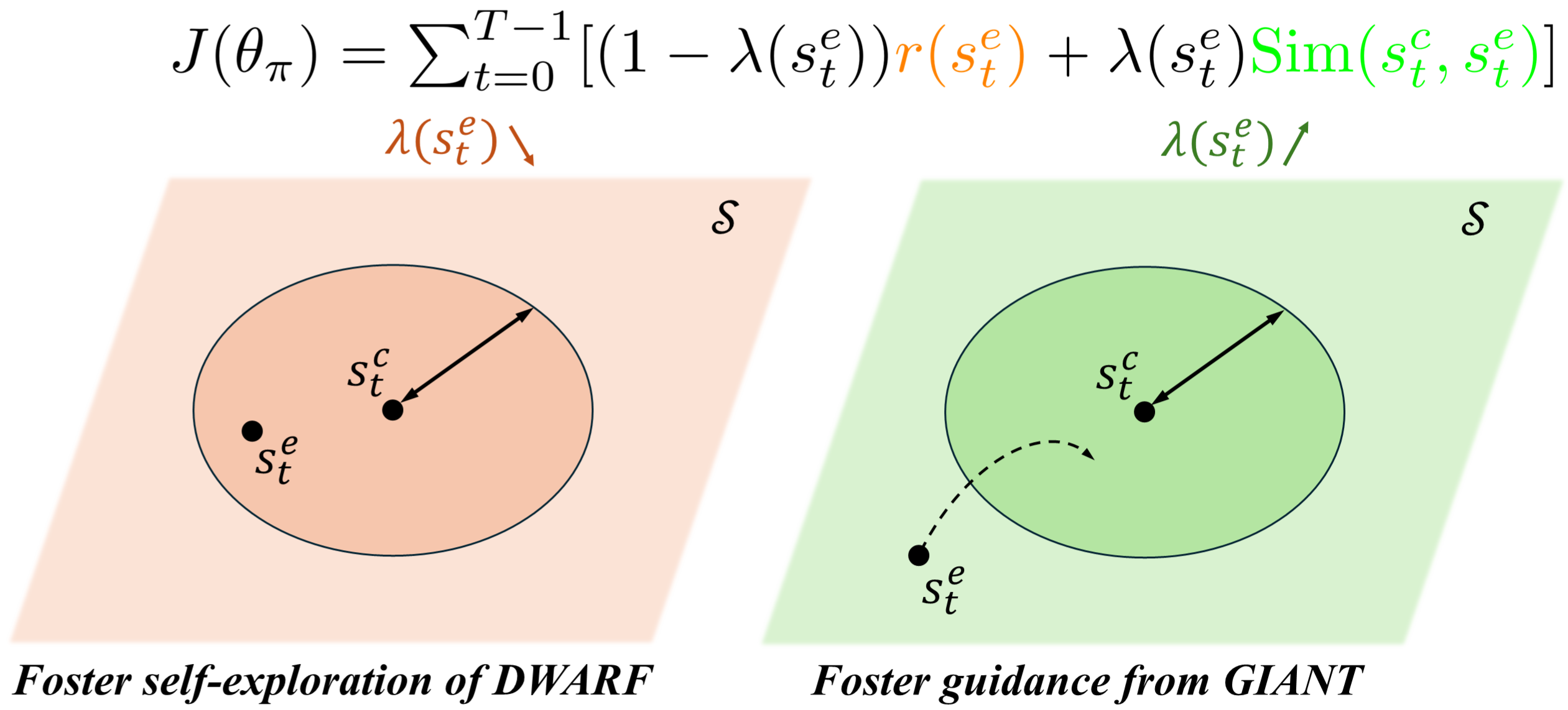}
    \caption{An illustration of Effective Guidance-Exploration. When DWARF is out of bounds, GIANT guides it to move quickly inside. Otherwise, DWARF prefers to explore for itself to find a correct target.}
    \label{fig:enter-label}
\end{figure}
Considering the two issues, we propose an efficient guidance-exploration approach, which gives DWARF a constraint reward to balance guidance and exploration via supervised learning approaches.\\
\textbf{Constraint Reward.} \quad Considering the constraint, that is, receiving high-quality guidance from GIANT, we introduce a metric for the state similarity between GIANT and DWARF, denoted as $\text{Sim}(s_t^c,s_t^e)$. It is calculated as the cosine similarity between the pre-trained embeddings of the current cluster\footnote{Each cluster embedding is obtained by averaging all entity embeddings within it. A slight abuse of notation indicates that the $\mathbf{c}_t$ here is not the $\mathbf{c}_t$ mentioned in the policy network. In the concrete implementation, we treat the $\mathbf{c}_t$ in the policy network as a self-cascade of the $\mathbf{c}_t$ here.} and the current entity: 
\begin{equation}
    \text{Sim}(s_t^c,s_t^e) = \frac{\mathbf{c}_t^\top \mathbf{e}_t}{||\mathbf{c}_t||_2 ||\mathbf{e}_t||_2}.
\end{equation}
We solve the optimization problem by the following formulas,
\begin{equation}
    \begin{aligned}
    & \text{max } \mathbb{E}_{a_1^e,...,a_T^e \sim \pi_\theta^e}\left[\sum_{t=0}^{T-1}r_e(s_t^e|s_0^e)\right] \\
    & \text{s.t. } \text{Sim}(s_t^c,s_t^e) \geq \frac{\delta}{r_c(s_{t}^c) + \varepsilon}
    \end{aligned}
\end{equation}
where $r_e(s_t^e)$ and $r_c(s_t^c)$ are default rewards for DWARF and GIANT respectively, the agent obtains a favorable reward 1 if the corresponding entity or cluster is a correct target and unfavorable reward 0 otherwise. It is only when GIANT reaches the correct cluster that the constraint in Equation 6 is operative. In practice, we set $\varepsilon = -0.01 \delta$.\\
\textbf{Practical Algorithm.} \quad The objective in Equation 6 can be optimized with any reinforcement learning algorithm that implements generalized policy iteration. Here we use REINFORCE \cite{DBLP:journals/ml/Williams92} and the method of Lagrange multipliers as described by \cite{DBLP:conf/iclr/AbdolmalekiSTMH18,DBLP:journals/corr/abs-2403-09930}. For all DWARF state $s_t^e$, we maximize the Lagrangian function, subject to $0 \leq \lambda(s_t^e) \leq 1$, 
\begin{equation} \label{eq7}
    J(\theta_{\pi^e}) = \sum_{t=0}^{T-1}[(1 - \lambda(s_t^e)) r_e(s_t^e) + \lambda(s_t^e) \text{Sim}(s_t^c,s_t^e)],
\end{equation}
the Lagrange multiplier is updated to make the guidance-exploration trade-off. Figure 3 gives an illustration: When the state similarity metric between GIANT and DWARF $\text{Sim}(s_t^c,s_t^e)$ is less than the threshold, the parameter $\theta_\lambda$ are optimized so that $\lambda(s_t^e)$ increases to encourage GIANT to provide more guidance for DWARF. Conversely, the parameter $\theta_\lambda$ are optimized so that $\lambda(s_t^e)$ decreases to encourage DWARF to explore in the constrictive space when the state similarity metric between GIANT and DWARF $\text{Sim}(s_t^c,s_t^e)$ is greater than the threshold. In practice, we utilize a cross-entropy loss to optimize $\theta_\lambda$:
\begin{equation} \label{eq8}
    \begin{aligned}
       & J(\theta_\lambda) = \sum_{t=0}^{T-1}[-(1-y)\text{log}(1-\lambda(s_t^e)) - y\text{log}(\lambda(s_t^e))] \\
       & \text{where }y = \left\{
            \begin{aligned}
                & 0 \quad \quad \text{if} \ \text{Sim}(s_t^c,s_t^e) \geq \frac{\delta}{r_c(s_{t}^c) + \varepsilon} \\
                & 1 \quad \quad \text{otherwise}
            \end{aligned}
       \right.
    \end{aligned}
\end{equation}

\subsection{Dynamic Path Feedback}
A further challenge is to enhance the search efficiency of GIANT in order to provide DWARF with the requisite guidance as expeditiously as possible, given that the search is limited to a fixed number of step $T$. In the default reward, GIANT will receive 1 reward only when it reaches the correct cluster, and the rest will be 0, which makes GIANT adopt random policy at the early phase. This phenomenon is not conducive to stable learning outcomes for two agents.
On the basis of the theory of reward shaping \cite{DBLP:conf/aaai/HarutyunyanDVN15}, we rewrite the reward function of GIANT, named dynamic path feedback. In particular, we write it in the form of an objective function $J(\theta_{\pi^c})$, 
\begin{equation} \label{eq9}
    \begin{aligned}
        & J(\theta_{\pi^c}) = \sum_{t=0}^{T-1}[ r_c(s_t^c) - \alpha 
          \Delta(s_t^c, s_{t+1}^c)], \\
        & \Delta(s_t^c, s_{t+1}^c) = \text{Sim}(s_t^c,s_{\text{target}}^c) - \text{Sim}(s_{t+1}^c,s_{\text{target}}^c)
    \end{aligned}
\end{equation}
$s_{t+1}^c$ comes from the next state generated by the policy network. In contrast to the default reward, dynamic path feedback uses the reward function to score the GIANT's path in a rollout rather than simply identifying whether it has reached the correct target cluster. Even if GIANT does not reach the correct target cluster in a rollout, it will evaluate the quality of the path, thereby accelerating the learning process. In Appendix D, we prove that GIANT learns optimal policy in dynamic path feedback is consistent with default rewards circumstance:
\begin{theorem}[\textit{Consistency of optimal policy}] \label{th1}
    Given two MDPs that differ only in reward function, denoted as $M = (\mathcal{S}, \mathcal{A}, \delta, \mathcal{R})$ and $M' = (\mathcal{S}, \mathcal{A}, \delta, \mathcal{R}_D)$ respectively, where $\mathcal{R} = r_c(s_t^c)$ is the default reward while $\mathcal{R}_D = r_c(s_t^c) - \alpha \Delta(s_t^c, s_{t+1}^c)$ is the dynamic path feedback reward. Their optimal policies are consistent, that is, 
    \begin{equation}
        \pi^*_{M}(s_t^c) = \pi^*_{M'}(s_t^c).
    \end{equation}
\end{theorem}

\section{Experiments}
In this section, we evaluate the efficacy of FULORA on three real-world KG datasets: NELL-995 \cite{DBLP:conf/emnlp/XiongHW17}, WN18RR \cite{DBLP:conf/aaai/DettmersMS018} and FB15K-237 \cite{DBLP:conf/emnlp/ToutanovaCPPCG15}. The datasets statistics are listed in Table 1.
\begin{table}[t] \label{table1}
    \centering
    \small
    \setlength{\tabcolsep}{0.75mm}
    \begin{tabular}{ccccccc}
    \toprule
     \multirow{2}{*}{\textbf{Dataset}} & \multirow{2}{*}{\textbf{\#Ent}} & \multirow{2}{*}{\textbf{\#Rel}} & \multirow{2}{*}{\textbf{\#Fact}} & \multirow{2}{*}{\textbf{\#Que}} & \multirow{2}{*}{$\textbf{\#Mean}$} & \multirow{2}{*}{$\textbf{\#Med}$} \\ \\
     \midrule
     NELL-995 & 75,492 & 200 & 154,213 & 3,992 & 4.07 & 1 \\
     WN18RR & 40,945 & 11 & 86,835 & 3,134 & 2.19 & 2 \\
     FB15K-237 & 14,505 & 237 & 272,115 & 20,466 & 19.74 & 14 \\
     \bottomrule
    \end{tabular}
    \caption{The statistics of some benchmark KG datasets. \#Mean is the averaged outgoing degree of every entity that can indicate the sparsity level while \#Med is the corresponding median.}
    \label{tab:my_label}
\end{table}

\begin{table*}[t] \label{table2}
\small
\setlength{\tabcolsep}{2.5mm}
\begin{tabular}{@{}l|cccc|cccc|cccc@{}}
\toprule
\multirow{2}{*}{\textbf{Model}} & \multicolumn{4}{c}{\textbf{NELL-995}} & \multicolumn{4}{c}{\textbf{WN18RR}} & \multicolumn{4}{c}{\textbf{FB15K-237}} \\ \cmidrule(l){2-13} 
& MRR   & @1    & @3    & @10  & MRR   & @1   & @3   & @10  & MRR   & @1    & @3    & @10   \\ \midrule
TransE \cite{DBLP:conf/nips/BordesUGWY13}  & 51.4  & 45.6  & 67.8  & 75.1 & 35.9  & 28.9 & 46.4 & 53.4 & 36.1  & 24.8  & 40.1  & 45.0  \\
DistMult \cite{DBLP:journals/corr/YangYHGD14a} & 68.0  & 61.0  & 73.3  & 79.5 & 43.3  & 41.0 & 44.1 & 47.5 & 37.0  & 27.5  & 41.7  & 56.8  \\
ComplEx \cite{DBLP:conf/icml/TrouillonWRGB16}  & 68.4  & 61.2  & 76.1  & 82.1 & 41.5  & 38.2 & 43.3 & 48.0 & 39.4  & 30.3  & 43.4  & 57.2  \\
LMKE \cite{DBLP:conf/ijcai/WangHLX22}  & \underline{74.6}  & 71.7  & 84.7  & 89.5 & \underline{53.6} & 45.1 & 66.0 & 79.4 & \underline{41.2}  & 31.8 & 46.2  & 56.9 \\
\midrule
NBFNet \cite{zhu2021neural} & 70.9 & 68.0 & 76.8 & 80.4 & \underline{48.1} & 44.9 & 49.8 & 58.3 & \underline{40.5} & 30.8 & 45.8 & 55.2 \\
RED-GNN \cite{zhang2022knowledge} & \underline{71.2} & 67.8 & - & 86.2 & 46.9 & 42.5 & - & 53.0 & 39.8 & 29.9 & - & 54.4 \\
\midrule
MINERVA \cite{DBLP:conf/iclr/DasDZVDKSM18} & 67.5  & 58.8  & 74.6  & 81.3 & 44.8  & 41.3 & 45.6 & 51.3 & 27.1  & 19.2  & 30.7  & 42.6  \\
AttnPath \cite{DBLP:conf/emnlp/WangLPM19} & 69.3  & 62.7  & 73.9  & 80.1 & 42.9  & 40.7 & 44.3 & 52.9 & 31.9  & 24.1  & 40.4  & 43.8  \\
SQUIRE \cite{DBLP:conf/emnlp/BaiLL0QDX22} & 71.1  & 68.2  & 81.4  & 87.2 & 48.2 & 45.0 & 51.4 & 59.7 & 35.0  & 26.5  & 41.7  & 50.3  \\
CURL \cite{DBLP:conf/aaai/ZhangY0L022}    & 70.8  & 66.7  & 78.6  & 84.3 & 46.0  & 42.9 & 47.1 & 52.3 & 30.6  & 23.9  & 38.1  & 50.9  \\
HMLS \cite{DBLP:journals/tkde/ZhengCWZYZ24} & 71.8  & 69.0  & 80.9  & 88.9 & 48.5  & 43.9 & 52.9 & 60.4 & \textbf{37.4}  & 28.0 & 42.1  & 52.5 \\
\rowcolor{gray!20}
\textbf{FULORA (Ours)}   & \textbf{72.5} & 69.4 & 79.7  & 89.2 & \textbf{49.1} & 45.6 & 50.7 & 59.2 & 36.4  & 27.1  & 41.9  & 51.3  \\ \bottomrule
\end{tabular}
\caption{Link prediction results with a \textbf{path length of 3} on the NELL, WordNet, and Freebase datasets (embedding-based and GNN-based reasoning do not inherently consider path length, so we exclude triples reachable by shorter paths for fair comparison). All metrics are multiplied by 100. The best score of embedding-based reasoning models, GNN-based models are \underline{underlined} while multi-hop reasoning models are in \textbf{bold}. Compared to other indicators, we specifically highlight the MRR to emphasize its significance.}
\end{table*}
As can be seen from the statistical indicators, these three datasets represent standard KG, sparse KG and Dense KG respectively. In our selection of the baseline, we are not only evaluating it against the state-of-the-art multi-hop reasoning methods, but also against with other embedding-based KG reasoning methods and GNN-based reasoning methods. FULORA and all baselines are implemented under the Pytorch framework and run on the NVIDIA 3080Ti GPU. All the results are the average of the results in five experiments. See Appendix A.1 for a brief introduction of each baseline.

\subsection{Short-distance Reasoning Ability}
For each triplet \( (e_s, r_q, e_o) \) in the test set, we convert it to a query \( (e_s, r_q, ? ) \) and use embedding-based or multi-hop models with a beam search width of 50 to rank the tail entities. Following \cite{DBLP:conf/nips/BordesUGWY13}, we evaluate using two metrics: (1) mean reciprocal rank (MRR) and (2) Hits@K, the proportion of correct tail entities ranked in the top K.

As shown in Table 2, we present the performance of FULORA and all baselines on NELL-995, WN18RR, and FB15K-237. On both the standard KG (NELL-995) and sparse KG (WN18RR), FULORA not only outperforms CURL, SQUIRE and HMLS currently regarded as the most powerful multi-hop reasoning models, but also significantly surpasses NBFNet and RED-GNN, both of which are recognized as advanced GNN-based reasoning algorithms. FULORA also achieves comparable performance to the best embedding-based model, LMKE. On FB15K-237, where 1-to-M relations dominate, multi-hop models often struggle with high-degree nodes, hindering correct entity retrieval. In contrast, FULORA’s attention mechanism focuses on the most relevant neighbors. Averaging across three datasets, FULORA improves MRR and Hits@1, 3, and 10 by 3.5\%, 2.9\%, 2.8\%, and 4.1\% over CURL. \textbf{Notably, Table 2 presents results based on a path length of 3, which does not capture FULORA’s exceptional performance in long-distance reasoning.} Subsequent experiments focus on the performance of FULORA’s components, GIANT and DWARF, in long-distance reasoning, with comparisons to other advanced models in Table 2. In fact prediction, FULORA outperforms other multi-hop baselines, as detailed in Appendix B.1.

\begin{table*}[t] \label{table3}
    \centering
    \setlength{\tabcolsep}{5mm}
    \begin{tabular}{l|ccc}
        \toprule
        \textbf{NELL-995/WN18RR/FB15K-237} & MRR & @1 & @10 \\
        \midrule
         FULORA + TransE \cite{DBLP:conf/nips/BordesUGWY13} & 72.5/49.1/36.4 & 69.4/45.6/27.1 & 89.2/57.2/51.3 \\
         FULORA + LMKE \cite{DBLP:conf/ijcai/WangHLX22} & 76.8/54.4/38.6 & 73.0/46.3/30.2 & 93.7/72.4/51.2 \\
         \bottomrule
    \end{tabular}
    \caption{Effects of various pre-embedding methods on the performance of FULORA.}
    \label{tab:my_label}
\end{table*}

\subsection{Long-distance Reasoning Ability}
Recall the motivation of FULORA, we tend to address the issue of multi-hop reasoning models being unable to infer correct answer due to the lack of short direct path by improving long-distance reasoning ability. This issue is significant on standard KG and sparse KG, so we conduct the following experiments on NELL-995 and WN18RR. For effective evaluation, we compare our model with MINERVA and CURL in NELL-995 and WN18RR, where we remove the most frequently-visited short paths found by the bi-directional search \cite{DBLP:conf/emnlp/XiongHW17,DBLP:conf/aaai/ZhangY0L022} inside KGs.

\begin{figure}[t] \label{fig4}
    \centering
    \includegraphics[width=1\linewidth]{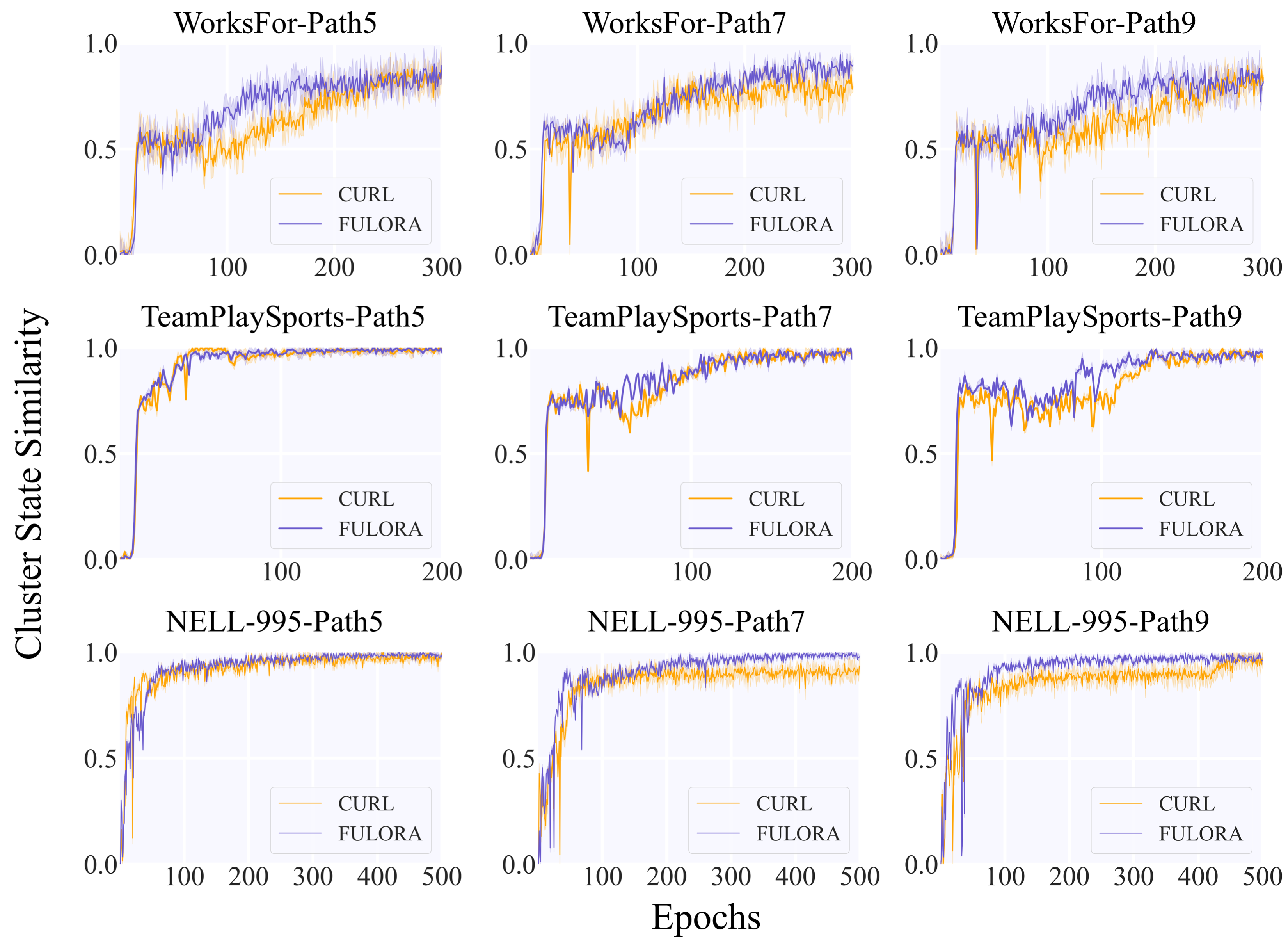}
    \caption{Learning curves comparing the performance of ours against CURL from NELL-995 relation tasks and all tasks. Averaged over 5 seeds with the shaded area showing standard deviation. Our proposed model is significantly better than CURL in both score and stability.}
    \label{fig:enter-label}
\end{figure}

\noindent
\textbf{Cluster State Similarity. \quad} One of the most significant contributions of GIANT is the implementation of dynamic path feedback, which has been shown to enhance the learning efficiency. To visually demonstrate the efficacy of dynamic path feedback, we initially focus on GIANT's reasoning ability on cluster-level, which directly impacts DWARF's reasoning results. Here, we record the cluster state similarity (CSS) $\text{Sim}(s_T^c,s_{\text{target}}^c)$ under each epoch. As shown in Figure 4, our model outperforms CURL in scores and stability in most cases. See Appendix B.3 for a comparison of the remaining tasks. 
\\
\textbf{Entity Reasoning Accuracy.} \quad we now demonstrate that FULORA's efficient guidance-exploration enhances DWARF’s performance in long-distance reasoning compared to other well-performed baselines (Appendix B.4.2 provides a further analysis). Figure 5 shows the MRR for varying path lengths on NELL-995 and WN18RR. FULORA excels in long-distance reasoning accuracy on WN18RR (sparse KG). This highlights our motivation: FULORA mitigates the poor performance of current KG reasoning methods in the absence of short direct paths by enhancing long-distance reasoning. Overall, FULORA demonstrates more robust long-distance reasoning performance with minimal degradation in long-path settings. This is due to the dynamic path feedback and efficient guidance-exploration, which enable better information sharing between GIANT and DWARF, ensuring DWARF’s excellent self-exploration ability.

\begin{figure}[t] \label{fig5}
    \centering
    \includegraphics[width=1\linewidth]{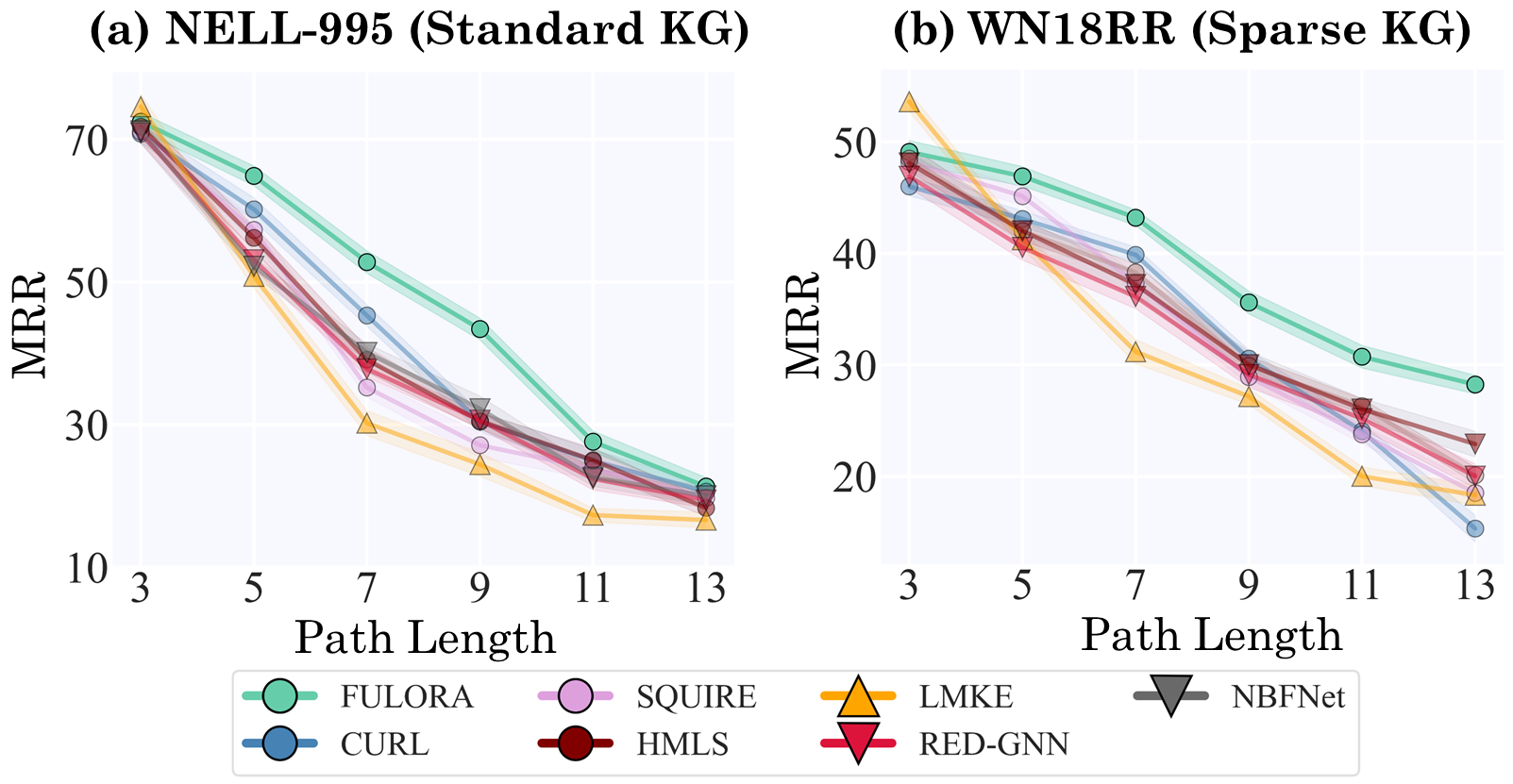}
    \caption{The long-distance performance: FULORA significantly outpeforms CURL, SQUIRE, HMLS, LMKE, RED-GNN, NBFNet on NELL-995 (standrad KG) and WN18RR (Sparse KG).}
    \label{}
\end{figure}

\subsection{Ablation Study}
\textbf{Sensitivity Analysis.} \quad We utilize $\delta$ and $\alpha$ to control the degree of efficient guidance-exploration and dynamic path feedback. To comprehensively explore the efficacy of the two mechanisms, we conduct link prediction on NELL-995 and WN18RR by varying the values of $\delta$ and $\alpha$ as $\{0.20,0.30,0.40,0.50\}$ and $\{0.05,0.10,0.15,0.20\}$, respectively. The averaged results are shown in Figure 6. In accordance with our previous analysis, it is evident that DWARF cannot rely excessively on GIANT for guidance. Furthermore, the level of path feedback that GIANT receives should not be unduly strong. In addition, in Appendix B.4, we have carefully discussed the influence of three main components of FULORA on KG reasoning efficiency.

\begin{figure} \label{fig6}
    \centering
    \includegraphics[width=1\linewidth]{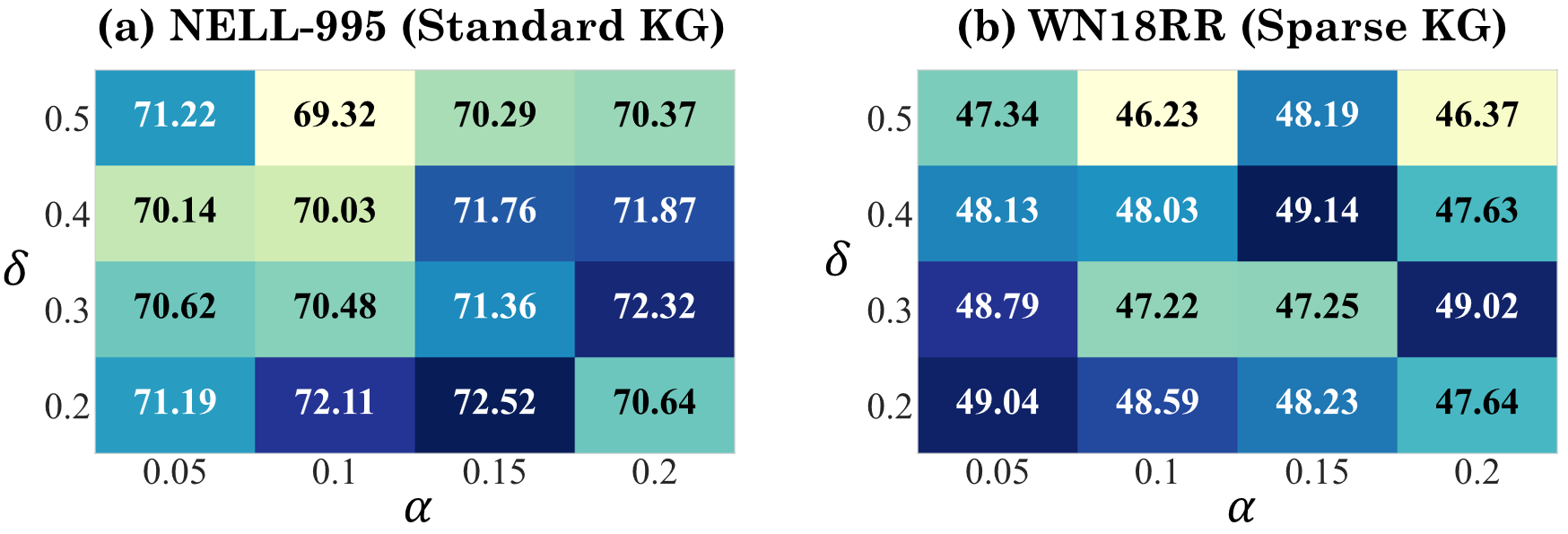}
    \caption{Ablation analysis w.r.t $\delta$ and $\alpha$ on NELL-995 and WN18RR, respectively. Here we report MRR results with path length=3 for both datasets.}
    \label{fig:enter-label}
\end{figure}

\noindent
\textbf{Pre-embedding Methods.} \quad In our approach, we use TransE \cite{DBLP:conf/nips/BordesUGWY13} for pre-training the embeddings. One straightforward idea is to improve FULORA's performance by using a more advanced KG embedding model such as LMKE \cite{DBLP:conf/ijcai/WangHLX22}. Table 3 presents the performance of TransE and LMKE as pre-embedding methods. The more advanced LMKE indeed improves FULORA’s performance. However, we use TransE in our experiments to emphasize that the performance improvement is due to FULORA’s efficient guidance-exploration mechanism, not the strength of the KG embedding method.

\section{Conclusion}
We present FULORA, an efficient guidance-exploration model built on dual-agent KG reasoning framework to enhance the agent's long-distance reasoning ability on standard KG and sparse KG. The key insight behind our approach is balancing the self-exploration of DWARF and the guidance from GIANT. Specifically, on the one hand, we leverage the attention mechanism to make DWARF pay attention to the neighbouring entities that are close to the query. On the other hand, we propose that dynamic path feedback enables GIANT to have better learning efficiency, thus providing DWARF with high-quality guidance, making the DWARF to have a favourable global vision while having excellent local reasoning ability. Experiments on three real-world datasets demonstrate that FULORA outperforms state-of-the-art multi-hop reasoning methods. Further analysis reveals that FULORA’s long-distance reasoning ability on standard and sparse KGs significantly outperforms current KG reasoning methods.

\clearpage

\appendix
\section{A \quad Experiment Setup}
\subsection{A.1 \quad Baseline Methods} \label{appendix a.1}
We compare the performance of FULORA with the following baselines, including embedding-based, GNN-based and multi-hop based KG reasoning methods:
\begin{enumerate}
    \item \textbf{TransE} \cite{DBLP:conf/nips/BordesUGWY13} embeds entities and relations into low-dimensional vector space so that complex relationships in graphs can be represented and reasoned by vector operations.
    \item \textbf{DistMult} \cite{DBLP:journals/corr/YangYHGD14a} uses triples (source entity, relation, tail entity) to train the model, where each entity and relation is represented as a vector. The target is to maximize the score of the correct triples while minimizing the score of the wrong triples.
    \item \textbf{ComplEx} \cite{DBLP:conf/icml/TrouillonWRGB16} is based on tensor decomposition and uses complex vectors to represent entities and relationships in order to capture complex relationships in the knowledge graph.
    \item \textbf{DeepPath} \cite{DBLP:conf/emnlp/XiongHW17} utilizes a knowledge graph-based embedded policy-based agent with continuous states, which extends its path in a KG vector space by sampling the most promising relationships. 
    \item \textbf{MINERVA} \cite{DBLP:conf/iclr/DasDZVDKSM18} formulates the query task as a reinforcement learning (RL) problem where the goal is to take the best sequence of decisions (choice of relation edges) to maximize the expected reward (reaching the correct answer node). 
    \item \textbf{AttnPath} \cite{DBLP:conf/emnlp/WangLPM19} makes use of attention mechanism to force an agent to walk forward every step to avoid the agent stalling at the same entity node constantly. 
    \item \textbf{NBFNet} \cite{zhu2021neural} parameterizes the generalized Bellman-Ford algorithm with 3 neural components, namely INDICATOR, MESSAGE and AGGREGATE functions, which corresponds to the boundary condition, multiplication operator, and summationoperator respectively
    \item \textbf{RED-GNN} \cite{zhang2022knowledge} makes use of dynamic programming to recursively encodes multiple r-digraphs with shared edges, and utilizes query-dependent attention mechanism to select the strongly correlated edges.
    \item \textbf{SQUIRE} \cite{DBLP:conf/emnlp/BaiLL0QDX22} utilizes an encoder-decoder Transformer structure to translate the query to a path, without relying on existing edges to generate the path.
    \item \textbf{LMKE} \cite{DBLP:conf/ijcai/WangHLX22} adopts Language Models to derive Knowledge Embeddings (KE) and formulate description-based KE learning with a contrastive learning framework to improve efficiency in training and evaluation. 
    \item \textbf{CURL} \cite{DBLP:conf/aaai/ZhangY0L022} trains two agents (GIANT and DWARF) to walk over a knowlegde graph jointly and search for the answer collaboratively.
    \item \textbf{HMLS} \cite{DBLP:journals/tkde/ZhengCWZYZ24} improves the generalizability and effectiveness of multi-hop reasoning in few-shot scenarios by exploiting hard relations and hierarchical relation structures. 
\end{enumerate}
\begin{table}[t] \label{table4}
    \centering
    \setlength{\tabcolsep}{1.6mm}
    \resizebox{1.0\linewidth}{!}{
    \begin{tabular}{l|ccc}
        \toprule
        Hyperparameter & NELL-995 & WN18RR & FB15K-237 \\
        \midrule
        Embedding size & 50 & 50 & 50 \\
        Hidden size & 50 & 50 & 50 \\
        Batch size & 128 & 256 & 512 \\
        Learning rate & 0.01 & 0.001 & 0.001 \\
        Optimizer & Adam & Adam & Adam \\
        Cluster number & 75 & 100 & 200 \\
        Beam search size & 100 & 50 & 100 \\
        $\alpha$ & 0.15 & 0.15 & 0.25 \\
        $\delta$ & 0.20 & 0.40 & 0.30 \\
        \bottomrule
    \end{tabular}}
    \caption{FULORA Hyperparameters on three KG datasets.}
    \label{tab:my_label}
\end{table}
\begin{table*}[t] \label{table5}
    \centering
    \resizebox{1.0\linewidth}{!}{
    \begin{tabular}{l|cc|cccccc}
        \toprule
        \textbf{Task} & \textbf{TransE} & \textbf{TransR} & \textbf{PRA} & \textbf{DeepPath} & \textbf{MINERVA} & \textbf{M-Walk} & \textbf{CURL} & \textbf{FULORA} \\
        \midrule
        PersonBornInLocation & 62.7 & 67.3 & 54.7 & 75.1 & 80.0 & \textbf{84.7} & 82.7 & 84.4 \\
        OrgHeadquarteredInCity & 62.0 & 65.7 & 81.1 & 79.0 & 94.0 & 94.3 & 94.8 & \textbf{95.1} \\
        AthletePlaysForTeam & 62.7 & 67.3 & 54.7 & 75.0 & 80.3 & 84.7 & 82.9 & \textbf{86.0} \\
        AthletePlaysInLeague & 77.3 & 91.2 & 84.1 & 96.0 & 94.2 & 96.1 & 97.1 & \textbf{97.3} \\
        AthletePlaysSport & 87.6 & 96.3 & 47.4 & 95.7 & 98.0 & 98.3 & 98.4 & \textbf{98.5} \\
        TeamPlaysSport & 76.1 & 81.4 & 79.1 & 73.8 & 88.0 & 88.4 & 88.7 & \textbf{90.2} \\
        WorksFor & 67.7 & 69.2 & 68.1 & 71.1 & 81.0 & 83.2 & 82.1 & \textbf{84.3} \\
        \bottomrule
    \end{tabular}}
    \caption{Fact prediction results on seven tasks from NELL-995. All metrics are multiplied by 100. The best score of all models is in \textbf{bold}.}
    \label{tab:my_label}
\end{table*}
\subsection{A.2 \quad Data Statistics} \label{appendix a.2}
We adopt three KG datasets with different scales: NELL-995 (standard KG), WN18RR (sparse KG) and FB15K-237 (dense KG). We give a brief overview of these datasets.

\textbf{NELL-995} \cite{DBLP:conf/emnlp/XiongHW17} is an open source machine learning dataset developed by the OpenAI research group that contains more than 950,000 pieces of entity relationship data collected from the network to help machine learning systems make inferences. NELL995 can be used to train machine learning models, such as natural language processing models, machine translation models, question answering systems, and semantic search systems. 

\textbf{WN18RR} \cite{DBLP:conf/aaai/DettmersMS018} is a subset of WordNet that describes the association characteristics between English words. It preserves the symmetry, asymmetry, and composition relationships of the WordNet, and removes the inversion relationships. WN18RR contains some relational information about words. It consists of 14,541 entities and 237 relationships.

\textbf{FB15K-237} \cite{DBLP:conf/emnlp/ToutanovaCPPCG15} is a common knowledge graph dataset, which is a subset extracted from Freebase knowledge graph. It contains 14,505 entities and 237 relationships, and the data is carefully processed to remove reversible relational data and trivial triples, ensuring that entities in the training set are not directly connected to the verification or test set, thus avoiding information leakage issues.

\subsection{A.3 \quad Implementation Details} \label{appendix a.3}
We use the following software versions:
\begin{itemize}
    \item Ubuntu 24.04 LTS
    \item Python 3.10
    \item Pytorch 2.0.1
\end{itemize}

We conduct all experiments with a single NVIDIA GeForce 3080Ti GPU. Our experiments of all of 12 baselines are conducted on their official implementation provided by their respective authors. To reproduce the results of our model in Table 2 and Table 3, we report the empirically optimal crucial hyperparameters as shown in Table 4. On all datasets, the quantities of path rollouts in training and testing are 20 and 100, separatively. The core codes of FULORA are available at: https://github.com/KotoHanon/LOCOCO.

\section{B \quad Additional Results and Analysis}
\subsection{B.1 \quad Fact Prediction Results} \label{appendix b.1}
As opposed to link prediction, fact prediction task is concerned with verifying the veracity of an unknown fact, the true test triplets are ranked with some generated false triplets. Since we share a similar query-answering mechanism as CURL \cite{DBLP:conf/aaai/ZhangY0L022}, FULORA is capable of identifying the most appropriate entity for a given query and eliminates the need to evaluate negative samples of any particular relation. In the experiments, the dual agents try to infer and walk through the cluster-level and entity-level respectively to reach the correct target under removing all links of groundtruth relations in the original KG. Here, we report Mean Average Precision (MAP) scores for various relation tasks of NELL-995. We reuse the results of TransE \cite{DBLP:conf/nips/BordesUGWY13}, TransR \cite{DBLP:conf/aaai/LinLSLZ15}, PRA \cite{DBLP:journals/ml/LaoC10}, DeepPath \cite{DBLP:conf/emnlp/XiongHW17}, MINERVA \cite{DBLP:conf/iclr/DasDZVDKSM18}, M-Walk \cite{DBLP:conf/nips/ShenCHGG18} on seven tasks already reported in \cite{DBLP:conf/aaai/ZhangY0L022}. As demonstrated in Table 5, FULORA produces a satisfying result in most tasks, contributing an average gain of 9.1\% relative to the multi-hop based reasoning approaches (PRA, DeepPath, MINERVA, M-Walk and CURL) and 16.1\% gain compared to the embedding-based approaches (TransE and TransR).

\subsection{B.2 \quad Case Stuides} \label{appendix b.2}
\begin{table*}[t] \label{table6}
    \centering
    \resizebox{1.0\linewidth}{!}{
    \begin{tabular}{l}
        \toprule
        \textbf{WorksFor (Answer):} \quad Jeff Skilling $\overset{worksfor}{\rightarrow}$ Enron \\
        \textbf{WorksFor (AttnPath):} \quad Jeff Skilling(2)
        $\overset{worksfor}{\rightarrow}$ Enron(2) $\overset{topmemberoforganizition}{\rightarrow}$ Kenneth Lay $\overset{personleadsorganization}{\rightarrow}$ \\ 
        Enron and Worldcom(5) \\
        \textbf{WorksFor (CURL):} \quad Jeff Skilling(2) $\overset{personleadsorganization}{\rightarrow}$ Enron and Worldcom \\ $\overset{subpartoforganization}{\rightarrow}$ GE $\overset{subpartoforganization^{-1}}{\rightarrow}$ Enron and Worldcom 
        $\overset{personleadsorganization^{-1}}{\rightarrow}$ Jeff Skilling \\$\overset{worksfor}{\rightarrow}$ Enron(3) \\
        \rowcolor{gray!20}
         \textbf{WorksFor (FULORA):} \quad Jeff Skilling(4) $\overset{worksfor}{\rightarrow}$ Enron $\overset{topmemberoforgnazition}{\rightarrow}$ Kenneth Lay(2) \\\rowcolor{gray!20} $\overset{worksfor}{\rightarrow}$ Enron(3) \\
        \midrule
        \textbf{AthletePlaysSport (Answer):} \quad Carlos Villanueva $\overset{athleteplayssport}{\rightarrow}$ Baseball \\
        \textbf{AthletePlaysSport (AttnPath):} \quad Carlos Villanueva(2) $\overset{athleteflyouttosportsteamposition}{\rightarrow}$ Center \\ $\overset{athleteflyouttosportsteamposition^{-1}}{\rightarrow}$ Chris Coste(3) $\overset{athleteplayssport}{\rightarrow}$ Baseball(2)\\
        \textbf{AthletePlaysSport (CURL):} \quad Carlos Villanueva(4) $\overset{athleteflyouttosportsteamposition}{\rightarrow}$ Center \\ $\overset{athleteflyouttosportsteamposition^{-1}}{\rightarrow}$ J.C. Boscan $\overset{athleteplayssport}{\rightarrow}$ Baseball(2) \\
         \rowcolor{gray!20}\textbf{AthletePlaysSport (FULORA):} \quad Carlos Villanueva(3) $\overset{athleteplayssport}{\rightarrow}$ Baseball(5) \\
        \midrule
        \textbf{TeamPlaysinLeague (Answer):} \quad Duke University $\overset{teamplaysinleague}{\rightarrow}$ International \\
        \textbf{TeamPlaysinLeague (AttnPath):} \quad Duke University $\overset{templayssport}{\rightarrow}$ Basketball $\overset{teamplayssport^{-1}}{\rightarrow}$ Boise State Broncos \\ $\overset{teamplaysinleague}{\rightarrow}$ NCAA \\
        \textbf{TeamPlaysinLeague (CURL):} \quad Duke University $\overset{teamplayssport}{\rightarrow}$ Basketball $\overset{teamplayssport^{-1}}{\rightarrow}$ Bucks \\ $\overset{teamplaysinleague}{\rightarrow}$ NBA \\
        \rowcolor{gray!20}\textbf{TeamPlaysinLeague (FULORA):} \quad Duke University $\overset{teamplayssport}{\rightarrow}$ Basketball $\overset{teamplayssport^{-1}}{\rightarrow}$ Old Dominion University \\ \rowcolor{gray!20}$\overset{teamplaysinleague}{\rightarrow}$ International \\
        \bottomrule
    \end{tabular}}
    \caption{Paths found by AttnPath, CURL and FULORA, respectively. In order to describe the self-ring succinctly, we denote that agent stay at the same entity continuously $n$ times as Entity($n$).}
    \label{tab:my_label}
\end{table*}
In this part, we take \texttt{WorksFor}, \texttt{AthletePlaysSport} and \texttt{TeamPlaysinLeague} from NELL-995, as examples, to analyze these paths found by AttnPath \cite{DBLP:conf/emnlp/WangLPM19}, CURL \cite{DBLP:conf/aaai/ZhangY0L022} and FULORA. In order to concisely demonstrate model performance across \textbf{varying path lengths} (\texttt{WorksFor}, \texttt{AthletePlaysSport}, and \texttt{TeamPlaysinLeague}), we assign respective path lengths of $\{9,\ 7,\ 3\}$ representing long-distance reasoning (LD), medium-distance reasoning (MD), and short-distance reasoning (SD). As depicted in Table 6 during LD reasoning (\texttt{WorksFor}), FULORA effectively identifies target entities even when following an incorrect path. Conversely, both AttnPath and CURL exhibit limitations. Specifically, AttnPath struggles with re-establishing correct paths while CURL faces challenges in precise reasoning at a granular level. Without effective guidance, AttnPath becomes entangled by multiple entities associated with identical relationships. For instance, in tasks involving \texttt{TeamPlaysinLeague}, AttnPath may identify the correct relationship but navigate towards an wrong entity. Similarly, CURL encounters analogous issues stemming from its inability to strike a proper balance between exploration and guidance. Simply put, CURL leads DWARF too closely along GIANT’s trajectory without fully benefiting DWARF on account of distributional deviations at finer levels, resulting in inadequate exploratory capabilities. 
\begin{figure*}[t] \label{fig7}
    \centering
    \includegraphics[width=1\linewidth]{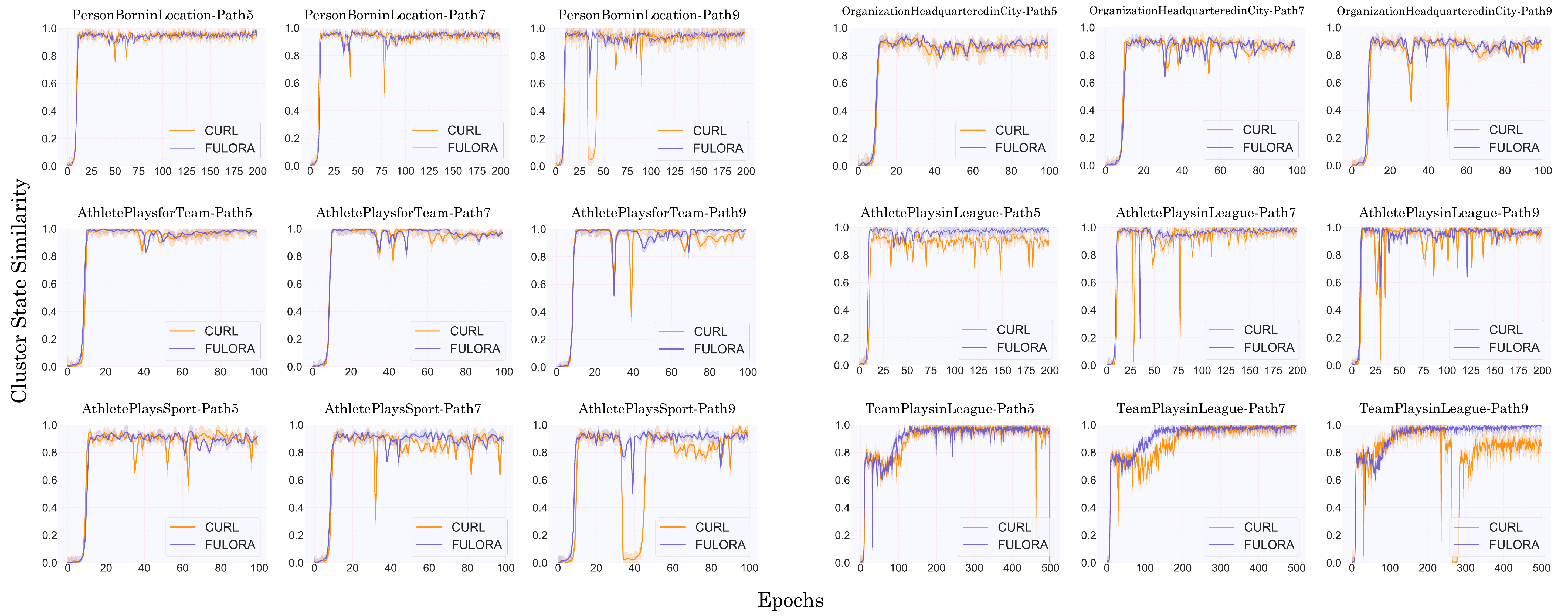}
    \caption{Learning curves comparing the performance of ours against CURL. Learning curves are averaged over 5 seeds, and the shaded area represents the standard deviation across seeds.}
    \label{fig:enter-label}
\end{figure*}

\subsection{B.3 \quad Cluster State Similarity} \label{appendix b.3}
To assess the exploration efficiency of GIANT, we introduce cluster state similarity (CSS) as a metric, and in the main content we compare the performance of CURL and FULORA on only three tasks due to space limitations. Figure 7 also compare the model performance on remaining six tasks. For the same task, CURL experiences significant oscillations as path length increases, while FULORA remains stable. As in our previous analysis, CURL can only judge path quality by whether it has gone to the correct target cluster, while FULORA utilizes dynamic path feedback to promote the GIANT to converge to a high-quality path.
\begin{figure}[t] \label{fig8}
    \centering
    \includegraphics[width=1\linewidth]{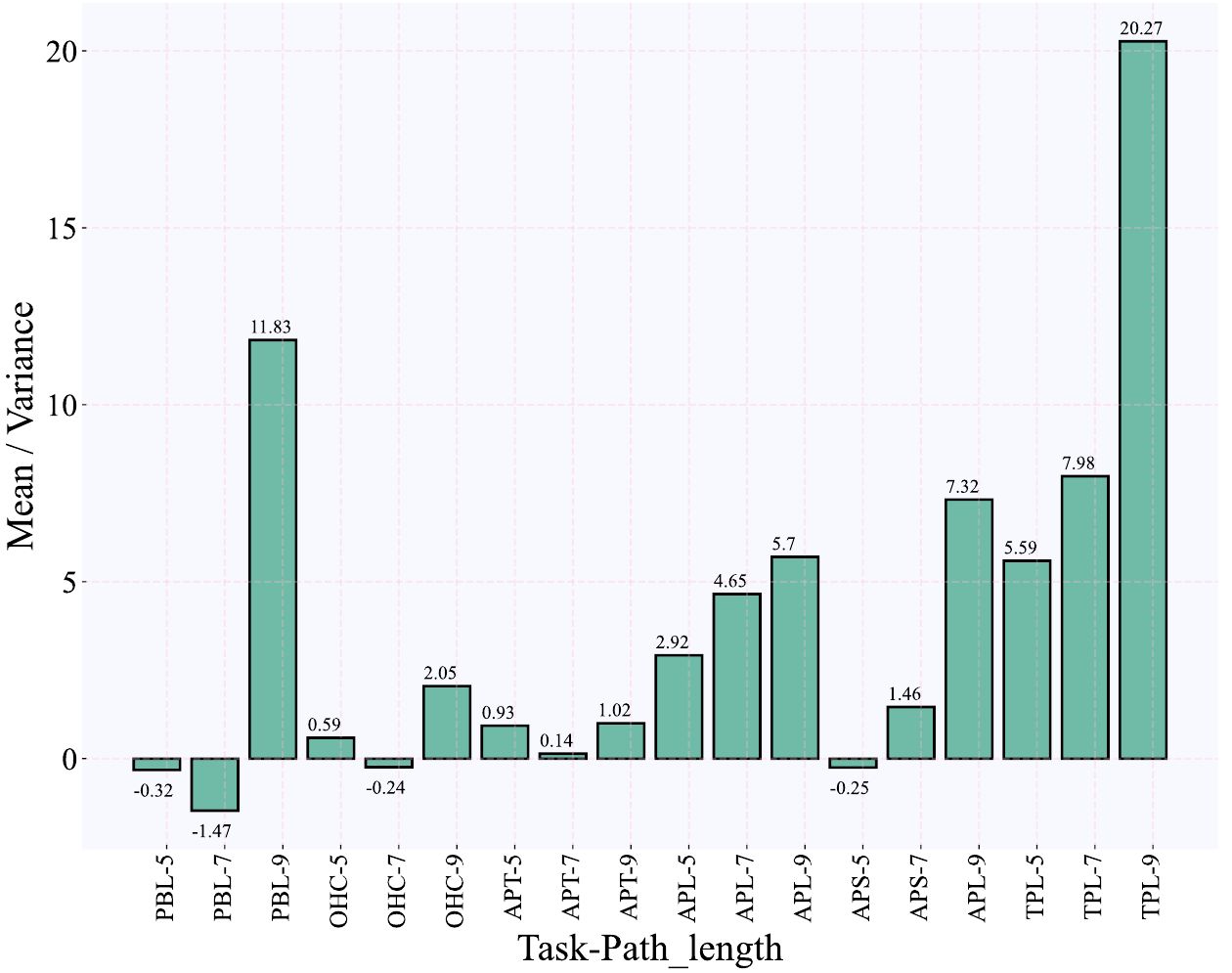}
    \caption{The ratios of the mean and variance of CSC difference between FULORA and CURL on different task-path\_length.}
    \label{fig:enter-label}
\end{figure}

Figure 8 further visualizes performances of FULORA and CURL on these tasks with varied path length. We utilize the ratio of the mean to the variance of CSS as an evaluation metric for the mean measures reasoning accuracy and the variance measures reasoning stability. In the vast majority of cases, FULORA outperforms CURL. In the case of long-distance reasoning for complex tasks like \texttt{PersonBorninLocation-Path9} (PBL-9), \texttt{AthletePlaysinLeague-Path7} (APL-9), \texttt{AthletePlaysSport-Path9} (APS-9), and \texttt{TeamPlaysinLeague} (TPL), the performance gap between FULORA and CURL is significant. 

\subsection{B.4 \quad Additional Ablation Studies} \label{appendix b.4}
    Here, we conduct a series of ablation experiments designed to answer the following three research questions: \\
    \textbf{RQ1: }Can \textbf{attention mechanism} enhance DWARF's reasoning ability? \\
    \textbf{RQ2: }Can \textbf{dynamic path feedback} accelerate DWARF'S learning speed via improving GIANT'S reasoning ability?
    \\
    \textbf{RQ3: }Can \textbf{efficient guidance-exploration method} make the guidance-exploration trade-off?

    With a slight abuse of abbreviation, we use ATTN, DPF and GE to denote attention mechanism, dynamic path feedback and the efficient guide-exploration method respectively. To highlight the effect of each component, we set FULORA as the baseline. In Figure 9, FULORA-X represents \textbf{pulling module X out of the FULORA framework}. We conduct experiments on three different scales of KG to make a fair comparison. It is worth noting that in order to describe DWARF reasoning accuracy in the whole training process, Entity State Similarity (ESS, be similar to CSS we mentioned above) is used as an evaluation index.
\subsubsection{B.4.1 \quad Attention Mechanism}
    We first focus on ablation experiments on the attention mechanism. As shown in Figure 9, the absence of the attention mechanism has little effect on WN18RR (sparse KG), while the effect of the attention mechanism increases with increasing density. In particular, on FB15K-237 (dense KG), the effect of the attention mechanism on the average ESS reaches $1/3$. It is consistent with the purpose of introducing the attention mechanism, that is, enhancing the reasoning ability of DWARF in the case of multi-neighbor and multi-relation. The reason lies in DWARF are forced to prioritize relations and neighbors that are highly correlated with the query relations, which plays an important role in reasoning on dense KG with complex relations, but because sparse KG neighbor relations are not sophisticated, the attention mechanism does not have a substantial effect.

\begin{figure*}[t] \label{fig9}
    \centering
    \includegraphics[width=1\linewidth]{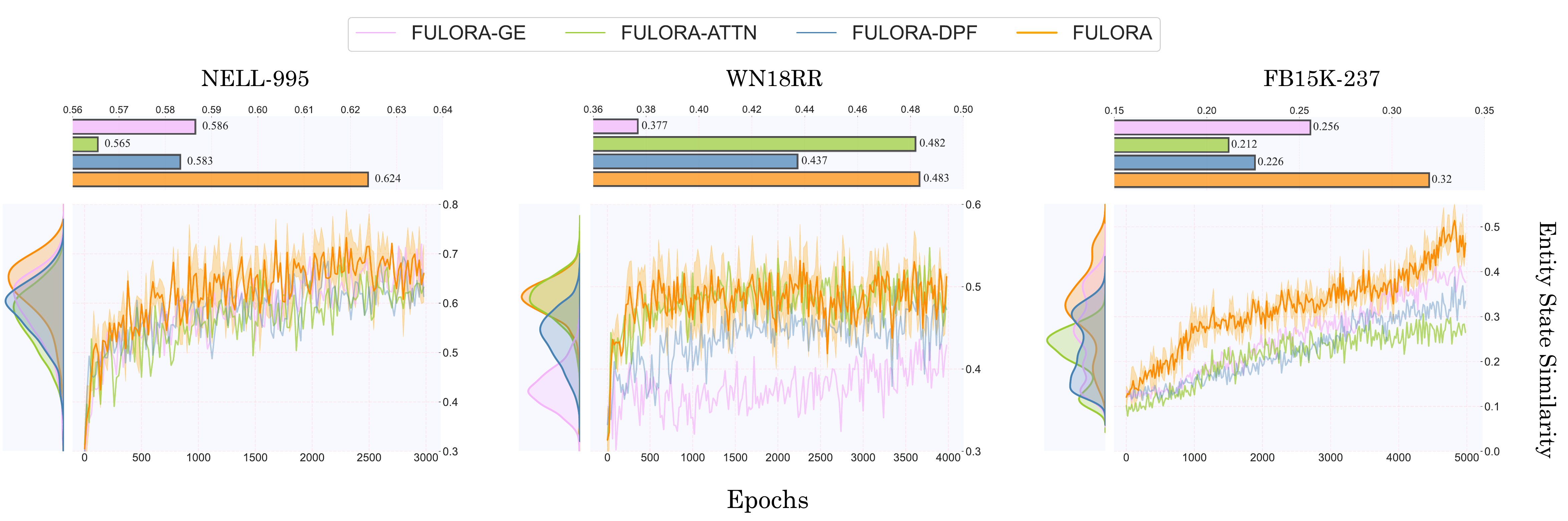}
    \caption{Entity state similarities comparing the performance of FULORA, FULORA-GE, FULORA-ATTN and FULORA-DPF. In addition, bar charts and distribution charts depict the learning speed and accuracy. Learning curves are averaged over 3 seeds.}
    \label{fig:enter-label}
\end{figure*}

\subsubsection{B.4.2 \quad Dynamic Path Feedback} \label{appendix b.4.2}
    We have demonstrated in previous experiments that dynamic path feedback can improve GIANT's reasoning ability, next we concentrate on the impact to DWARF's learning speed of dynamic path feedback. To illustrate the speed and stability of the training, we utilize kdeplot in Figure 9. The distribution maps on the three datasets have a common feature: compared with FULORA, the learning speed and learning accuracy decrease in the absence of dynamic path feedback (corresponding to the widening and downward movement of the distribution maps). In FB15K-237 (dense KG), the lack of dynamic path feedback has a great impact on the learning speed and accuracy, because the error of cluster mapping escalates with the complexity of neighbor relations when the number of clusters maintains. At this time, a large deviation between cluster-level KG and entity-level KG requires GIANT to make a correct evaluation of the path timely to provide exact guidance for DWARF.

\subsubsection{B.4.3 \quad Efficient Guidance-Exploration} \label{appendix b.4.3}
    The core of FULORA is the efficient guidance-exploration method, which contributes to make the guidance-exploration trade-off. In the previous Hierarchical RL-based reasoning method like CURL, there is a high degree of coupling between the two agents. \textbf{We treat ESS and CSS as two time series because they change with training progress}. Next, we use Granger Causality test \cite{diks2006new} to examine the degree of coupling between DWARF and GIANT.

    Granger Causality test is used to study the causal relationship between two sets of data, that is, to test whether one set of time series causes changes in another set of time series \cite{diks2006new}. It is worth noting that Granger Causality test requires stationary time series, otherwise false regression issues may occur, so it is necessary to detect the stationarity of time series through ADF test. If the pair-to-pair time series is non-stationary and satisfies the homogeneity of order, the Granger Causality test can be carried out only after the cointegration test between pair-to-pair sequences exists \cite{dougherty2011introduction}. Our time series analysis for ESS and CSS are detailed in Appendix C.

    The results of the Granger Causality Test for FULORA and FULORA-GE are shown in Table 7. The efficient guidance-exploration method reduces the coupling between GIANT and DWARF, thus alleviating the affect of policy shift caused by cluster mapping. In particular, on the WN18RR and FB15K-237, FULORA performs better than the version without efficient guidance-exploration method due to the greatly reduced coupling between DWARF and GIANT. Certain tasks in WN18RR (sparse KG) require the agent to have excellent long-distance reasoning ability. However, due to the distribution deviation, even if the GIANT reaches the correct target cluster, the guidance provided by it may not be suitable for DWARF (for example, the path length is too long to reach the correct target entity). The high coupling also seriously affects GIANT, resulting in poor model performance. 
\begin{figure*}[t] \label{fig10}
    \centering
    \includegraphics[width=1\linewidth]{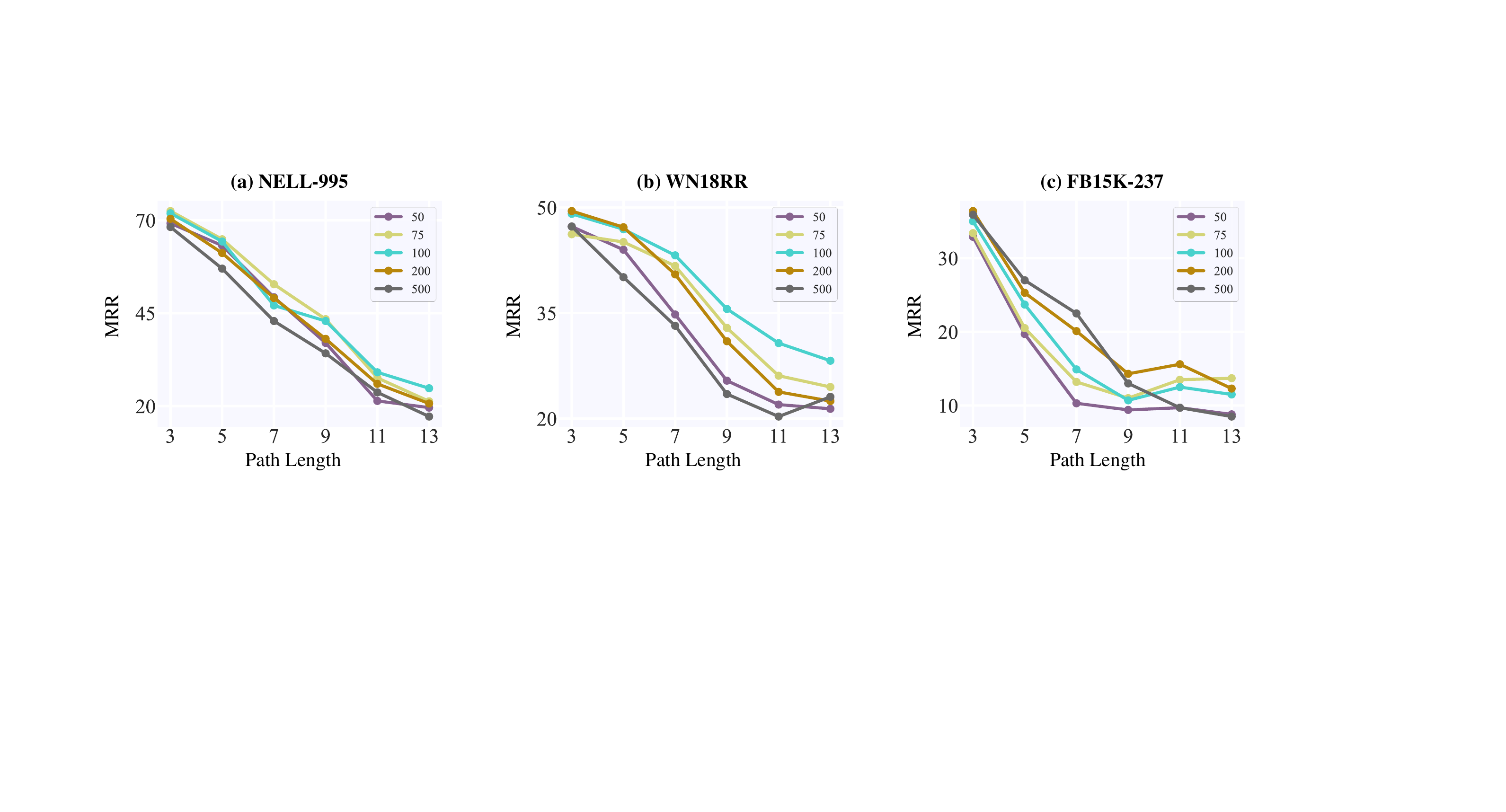}
    \caption{The effect of different cluster size used by GIANT. We present the link prediction (query answering) performance on three real-world KG datasets.}
    \label{fig:enter-label}
\end{figure*}

\begin{table}[t] \label{table7}
\setlength{\tabcolsep}{1mm}
\resizebox{1.0\linewidth}{!}{
\begin{tabular}{l|ccc}
\toprule
\textbf{Relation} & \textbf{NELL-995} & \textbf{WN18RR} & \textbf{FB15K-237} \\
\midrule
DWARF $\rightarrow$ GIANT & 21.5 & 9.6 & 18.9\\

DWARF(-GE) $\rightarrow$ GIANT(-GE) & 25.3 & 33.9 & 60.4 \\
\midrule
GIANT $\rightarrow$ DWARF & 21.7 & 40.3 & 36.3 \\

GIANT(-GE) $\rightarrow$ DWARF(-GE) & 26.3 & 41.1 & 37.3 \\
\bottomrule
\end{tabular}}
\caption{F statistics of Granger Causality test for FULORA and FULORA-GE. DWARF (-GE) and GIANT (-GE) are from FULORA-GE. In the Granger Causality test, a large F statistic indicates a strong causal relation. We set the lagged step as 2.}
\end{table}

\subsection{C \quad Sensitive Test on Cluster Size}
In the previous experiments, we examine the primary components of FULORA and analyze their impact on reasoning ability. Given that GIANT walks directly on the cluster-level KG derived from the entity-level KG, it is essential to separately evaluate the implications of cluster size. 

Figure 10 shows the MRR score of FULORA under different cluster number $N$ during training on three real-world KG datasets.  We observe that variations in cluster sizes significantly influence the reasoning performance of FULORA, particularly when reasoning over long distances on WN18RR (sparse KG) and FB15K-237 (dense KG).  If the cluster size is too small (e.g., $N = 500$), GIANT must traverse a greater number of clusters.  In the absence of higher-level agent to provide guidance, the reasoning process approximates the long-distance reasoning of a single agent. Conversely, when the cluster size is too large (e.g., $N = 50$), the guidance offered by GIANT becomes overly general. Even if the guidance is accurate, it may still prove ineffective, causing DWARF to similarly approximate the behavior of a single agent during long-distance reasoning.

\section{D \quad Time Series Analysis to Efficient Guidance-Exploration} \label{appendix c}
    Before Granger Causality test, we firstly use Augmented Dickey-Fuller (ADF) test \cite{cheung1995lag} to judge the stationarity of the series. The results as shown in Table 8. Next, we make D-order difference for the non-stationary time series, and then test the stationarity of the difference series. As illustrated in Figure 11, the time series is stationary after 1-order difference. Therefore, we demonstrate that these series are \textbf{integrated of order one}. To avoid the pseudo-regression phenomenon, we also need to do cointegration analysis of these series \cite{dougherty2011introduction}. Here, we employ Johansen test \cite{ho1996finding} based on maximum likelihood estimation for cointegration analysis.

    \begin{figure*}[h] \label{fig11}
    \centering
    \includegraphics[width=1\linewidth]{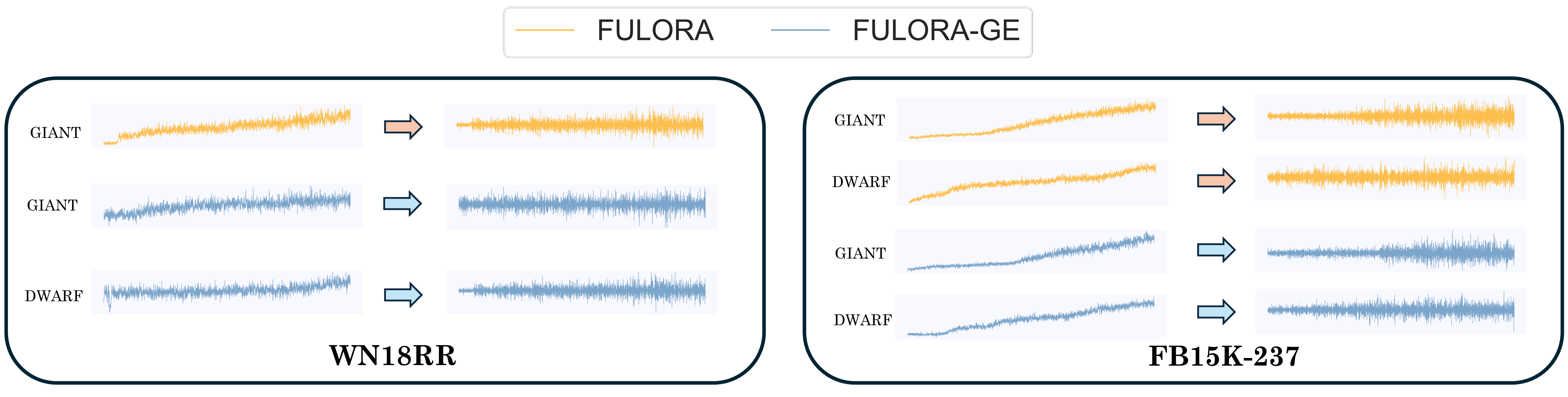}
    \caption{Entity state similarity curves and their 1-order difference on WN18RR and FB15K-237 from FULORA and FULORA-GE.}
    \label{fig:enter-label}
\end{figure*}

In the Johansen cointegration test \cite{ho1996finding}, the 2-cointegration term critical values of \{10\%, 5\%, 1\%\} are \{2.705, 3.841, 6.635\} respectively. The critical value 3.841 corresponding to 5\% was selected as the boundary for testing. Table 9 shows the results of Johansen cointegration test for FULORA and FULORA-GE. It shows that the seven non-stationary series avoid the pseudo-regression problem, thus the result of Granger causality test is reasonable (which is shown in Appendix B.4.3).

\begin{table}[t] \label{table8}
\setlength{\tabcolsep}{1.9mm}
\resizebox{1.0\linewidth}{!}{
\begin{tabular}{@{}l|cc|cc|cc@{}}
\toprule
\multirow{2}{*}{\textbf{Variable}} & \multicolumn{2}{c}{\textbf{NELL-995}} & \multicolumn{2}{c}{\textbf{WN18RR}} & \multicolumn{2}{c}{\textbf{FB15K-237}} \\ \cmidrule(l){2-7} 
& $t$   & $P$  & $t$  & $P$  & $t$  & $P$  \\ \midrule
DWARF & -3.25 & 0.04 & -7.40 & 0.00 & -1.67 &  \textbf{0.45} \\
GIANT & -5.22 & 0.00 & -1.46 & \textbf{0.46} & 0.08 & \textbf{0.97} \\
\midrule
DWARF(-GE) & -3.11 & 0.04 & -1.65 & \textbf{0.46} & 0.08 & \textbf{0.99} \\
GIANT(-GE) & -4.14 & 0.00 & -1.98 & \textbf{0.30} & -0.21 & \textbf{0.94} \\
\bottomrule
\end{tabular}}
\caption{t statistics and P value of ADF test for FULORA and FULORA-GE. DWARF (-GE) and GIANT (-GE) are from FULORA-GE. We set the bound of P value to 0.05, and P-values above the bound are \textbf{bolded} to indicate that they are not stationary series.}
\end{table}

\begin{table}[t] \label{table9}
\setlength{\tabcolsep}{1mm}
\resizebox{1.0\linewidth}{!}{
\begin{tabular}{@{}l|cc|cc@{}}
\toprule
\multirow{2}{*}{\textbf{Relation}} & \multicolumn{2}{c}{\textbf{WN18RR}} & \multicolumn{2}{c}{\textbf{FB15K-237}} \\ \cmidrule(l){2-5} 
& CR  & Trace  & CR  & Trace  \\ \midrule 
DWARF $\leftrightarrow$ GIANT & 0.019 & \textbf{38.436} & 0.002 & \textbf{4.736} \\

DWARF(-GE) $\leftrightarrow$ GIANT(-GE) & 0.037 & \textbf{74.582} & 0.001 & \textbf{4.133} \\
\bottomrule
\end{tabular}}
\caption{Results of Johansen cointegration test. CR is the abbreviation of Characteristic Root. The trace satisfying the critical value is \textbf{bolded}, indicating that there are two cointegration relations.}
\end{table}

\section{E \quad Proof of Dynamic Path Feedback} \label{appendix d}
    A major concern with dynamic path feedback is whether the optimal policy GIANT learns is consistent with the default rewards. Here we provide the following proof of Theorem 1.
    \begin{proof}
    Recall Equation 9 where the next state $s^c_{t+1}$ is used, which means that it implies the action $a^c_t$. The next state $s^c_{t+1}$ is the output of the current state $s^c_t$ and the current action $a^c_t$ to the transition function. In fact, $J(\theta_{\pi^c})$ is a function of state $s^c_t$ and action $a^c_t$. The same goes for reward settings in the default environment. Therefore, we use the Q-function to simplify our notation,
    \begin{equation}
        \begin{aligned}
            & Q(s^c_t,a^c_t) = \sum_{\tau = t}^{T-1} [r_c(s_\tau^c)].
        \end{aligned}
    \end{equation}

    The optimal Q-function $Q^*(s^c_t,a^c_t)$ is subject to Bellman optimal equation \cite{bellman1958dynamic}:
    \begin{equation}
        \begin{aligned}
            & Q^*(s_t^c,a_t^c) \\
            & = \mathbb{E}_{s_{t+1}^c} \Big[ r_c(s_t^c) + \underset{a_{t+1}^c \in \mathcal{A}}{\rm {max}} Q^*(s^c_{t+1},a^c_{t+1}) \Big],
        \end{aligned}
    \end{equation}
    then we make a simple transformation of the above formula to get
    \begin{equation}
        \begin{aligned}
            & Q^*(s_t^c,a_t^c) - \alpha \rm{Sim}(s^c_t,s^c_{\rm{target}}) \\
            & = \mathbb{E}_{s_{t+1}^c} \Big[ r_c(s_t^c) - \alpha \Delta(s_t^c, s_{t+1}^c) \\
            & + \underset{a_{t+1}^c \in \mathcal{A}}{\rm{max}} (Q^*(s^c_{t+1},a^c_{t+1}) - \alpha \rm{Sim}(s^c_{t+1},s^c_{\rm{target}})) \Big].
        \end{aligned}
    \end{equation}
    The $\rm{Sim}(s^c_t,s^c_{\rm{target}})$ we define is only related to the state, so Equation 12 is equivalent to Equation 13. Notice that $r_c(s_t^c) - \alpha \Delta(s_t^c, s_{t+1}^c)$ is the new reward function $\hat{r}_c(s^c_t)$ we designed for GIANT. Hence, Equation 12 can be viewed as
    \begin{equation}
        \begin{aligned}
            & \hat{Q}^*(s_t^c,a_t^c) \\
            & = \mathbb{E}_{s_{t+1}^c} \Big[ \hat{r}_c(s^c_t) + \underset{a_{t+1}^c \in \mathcal{A}}{\rm {max}} \hat{Q}^*(s^c_{t+1},a^c_{t+1}) \Big],
        \end{aligned}
    \end{equation}
    where $ \hat{Q}^*(s^c_t,a^c_t) = Q^*(s_t^c,a_t^c) - \alpha  \text{Sim}(s^c_t,s^c_{\rm{target}})$. Equation 14 is the Bellman optimal equation in dynamic path feedback. It constructs a new Markov Decision Process (MDP), which we use $M'$ to denote for distinction and $M$ for the original MDP. Thus, the optimal policy in $M'$ is 
    \begin{equation}
        \begin{aligned}
            \pi^*_{M'}(s_t^c) & = \rm{arg} \underset{a_{t}^c \in \mathcal{A}}{\rm{max}} \hat{Q}^*(s_t^c,a_t^c) \\
            & = \rm{arg} \underset{a_{t}^c \in \mathcal{A}}{\rm{max}} Q^*(s_t^c,a_t^c) - \alpha \rm{Sim}(s^c_t,s^c_{\rm{target}}) \\
            & = \rm{arg}\underset{a_{t}^c \in \mathcal{A}}{\rm{max}} Q^*(s_t^c,a_t^c) = \pi^*_{M}(s_t^c),
        \end{aligned}
    \end{equation}
     which implies that GIANT learns optimal policy in dynamic path feedback is consistent with optimal policy in the default rewards.  $\hfill\square$
\end{proof}

\section{F \quad Pseudocode for FULORA} \label{appendix e}
We show FULORA's training process in one episode.
\begin{algorithm}[h]
\caption{FULORA Training Algorithm (one episode)} 
\label{alg:Framwork} 
\begin{algorithmic}[1]
\REQUIRE ~~\\ %Input
Entity-level KG $\mathcal{G}^e$ and cluster-level KG $\mathcal{G}^c$; Initial policy networks parameters $\theta_{\pi^e}$ and $\theta_{\pi^c}$; Initial Lagrange multiplier parameter $\theta_\lambda$ ;Source entity and cluster nodes $e_s$ and $c_s$; Entity-level query $r_q$; Target entity and cluster nodes $e_o$ and $c_o$; Maximum path length $T$
\ENSURE ~~\\ %Output
parameters $\theta_{\pi^e}$, $\theta_{\pi^c}$, $\theta_\lambda$
\FOR{$t = 0,...,T-1$}
\STATE Set default cluster-level reward $r_c = 1$ if $c_t = c_o$ otherwise $r_c = 0$
\STATE Set default entity-level reward $r_e = 1$ if $e_t = e_o$ otherwise $r_e = 0$
\STATE Predict the action $a_t^c$ and $a_t^e$ for GIANT and DWARF based on policy networks parameters $\theta_{\pi^c}$ and $\theta_{\pi^e}$
\STATE Compute $J(\theta_{\pi^c})$, $J(\theta_{\pi^e})$, $J(\theta_\lambda)$ based on Equation 7-9.
\ENDFOR
\STATE Update model parameters: \\
$ \theta_{\pi^c} \leftarrow  \theta_{\pi^c} + 
      \alpha_{\pi^c} \nabla_{\theta_{\pi^c}} J(\theta_{\pi^c})$ \\
$ \theta_{\pi^e} \leftarrow  \theta_{\pi^e} + 
      \alpha_{\pi^e} \nabla_{\theta_{\pi^e}} J(\theta_{\pi^e})$ \\
$ \theta_{\lambda} \leftarrow  \theta_{\lambda} - 
      \alpha_{\lambda} \nabla_{\theta_{\lambda}} J(\theta_{\lambda})$ \\
\RETURN $\theta_{\pi^e}$, $\theta_{\pi^c}$, $\theta_{\lambda}$
\end{algorithmic}
\end{algorithm}

In addtion, a common concern is how clusters are formed, therefore, we provide pseudocode to demonstrate cluster formation, which is presented below.
\begin{algorithm}[h]
\caption{Cluster Formation and Batch Processing} 
\label{alg:Framwork} 
\begin{algorithmic}[1]
\REQUIRE ~~\\ Raw KG $\mathcal{G}$; Input data file, vocabularies (\texttt{entity\_vocab}, \texttt{relation\_vocab}, \texttt{cluster\_vocab}), batch size \texttt{batch\_size}, entity-to-cluster mapping \texttt{entity\_id\_to\_cluster\_mapping}
\ENSURE ~~\\ Generated batches with cluster-level relations.
\STATE Initialize vocabularies and mappings from the input data file
\STATE Parse the input file to extract triples ($e_1$,$r$,$e_2$)
\STATE Map each entity to its corresponding cluster
\STATE Create cluster relations $c_r = c_1 \_ c_2$ for each triple 
\STATE Store the triples and update entity-level and cluster-level mappings
\STATE For each batch, randomly sample triples from the stored data
\STATE Extract entities $e_1,e_2$, relations $r$ and their cluster mappings $c_1,c_2$
\STATE Yield entity-level and cluster-level batch data for training or testing
\RETURN Entity-level and cluster-level batch data.
\end{algorithmic}
\end{algorithm}

\end{document}